\documentclass[10pt,twocolumn,letterpaper]{article}

\usepackage[pagenumbers]{cvpr} 

\usepackage{graphicx}
\usepackage{amsmath}
\usepackage{amssymb}
\usepackage{booktabs}
\usepackage{epsfig}
\usepackage{graphicx}
\usepackage{bm}
\usepackage{makecell}
\usepackage{enumitem}
\usepackage{multirow}
\makeatletter
\@namedef{ver@everyshi.sty}{}
\makeatother
\usepackage{pgfplots}
\usepackage{booktabs}
\usepackage{multirow}
\usepackage[normalem]{ulem}
\usepackage{ifsym}

\newcommand\blfootnote[1]{%
\begingroup
\renewcommand\thefootnote{}\footnote{#1}%
\addtocounter{footnote}{-1}%
\endgroup
}

\usepackage[pagebackref,breaklinks,colorlinks]{hyperref}

\usepackage[capitalize]{cleveref}
\crefname{section}{Sec.}{Secs.}
\Crefname{section}{Section}{Sections}
\Crefname{table}{Table}{Tables}
\crefname{table}{Tab.}{Tabs.}

\definecolor{AMR}{rgb}{0.5,0.5,1.}

\definecolor{MS}{rgb}{1.0,0.5,0.0}

\definecolor{MANM}{rgb}{0.,0.5,1.0}

\definecolor{ben}{rgb}{0.9,0.,0.5}

\definecolor{todo}{rgb}{1.0, 0., 0.}

\def\cvprPaperID{9391} 
\def\confName{CVPR}
\def\confYear{2022}

\begin{document}

\title{3D-VField: Adversarial Augmentation of Point Clouds\\for Domain Generalization in 3D Object Detection}

\author{
Alexander Lehner$^{*,\circ,1,2}$
\quad Stefano Gasperini$^{*,1,2}$
\quad Alvaro Marcos-Ramiro$^{2}$
\quad Michael Schmidt$^{2}$\\
      Mohammad-Ali Nikouei Mahani$^{2}$
\quad Nassir Navab$^{1,3}$
\quad Benjamin Busam$^{1}$
\quad Federico Tombari$^{1,4}$\\\\
$^1$ Technical University of Munich \quad $^2$ BMW Group \quad
$^3$ Johns Hopkins University \quad $^4$ Google
}
\maketitle

\blfootnote{$^{*}$ The authors contributed equally.}
\blfootnote{$^{\circ}$ Contact author: Alexander Lehner (\textit{alexander.lehner@tum.de}).}
\blfootnote{Work partly sponsored by the German Federal Ministry for Economic Affairs and Energy (grant 19A19005B), VDA KI-Absicherung project.}


\begin{abstract}
As 3D object detection on point clouds relies on the geometrical relationships between the points, non-standard object shapes can hinder a method's detection capability. However, in safety-critical settings, robustness to out-of-domain and long-tail samples is fundamental to circumvent dangerous issues, such as the misdetection of damaged or rare cars. In this work, we substantially improve the generalization of 3D object detectors to out-of-domain data by deforming point clouds during training. We achieve this with 3D-VField: a novel data augmentation method that plausibly deforms objects via vector fields learned in an adversarial fashion. Our approach constrains 3D points to slide along their sensor view rays while neither adding nor removing any of them. The obtained vectors are transferable, sample-independent and preserve shape and occlusions. Despite training only on a standard dataset, such as KITTI, augmenting with our vector fields significantly improves the generalization to differently shaped objects and scenes. Towards this end, we propose and share CrashD: a synthetic dataset of realistic damaged and rare cars, with a variety of crash scenarios. Extensive experiments on KITTI, Waymo, our CrashD and SUN RGB-D show the generalizability of our techniques to out-of-domain data, different models and sensors, namely LiDAR and ToF cameras, for both indoor and outdoor scenes.
Our CrashD dataset is available at \href{https://crashd-cars.github.io}{https://crashd-cars.github.io}.
\end{abstract}

\section{Introduction}

With the established wide-spread progress of learning-based methods tackling a variety of perception tasks (e.g., object detection, semantic and panoptic segmentation), a recent trend denoted a focus shift towards ensuring the safe applicability of these powerful approaches in critical scenarios, such as autonomous driving and robotics~\cite{rabe2021development_safety}.
This has led to the pursuit of improving the model robustness
and generalization~\cite{mok2021advrush,gasperini2021r4dyn,wang2021generalizing}, especially against out-of-domain data, which can naturally occur in the real world~\cite{hendrycks2021natural}. Such approaches include domain adaptation~\cite{wang2020train} and generalization~\cite{wang2021generalizing}, uncertainty estimation~\cite{gasperini2021certainnet}, simulations~\cite{beery2020synthetic}, and adversarial alterations~\cite{tu_physically_2020}.

\begin{figure}[t]
\begin{center}
\includegraphics[width=1.00\linewidth]{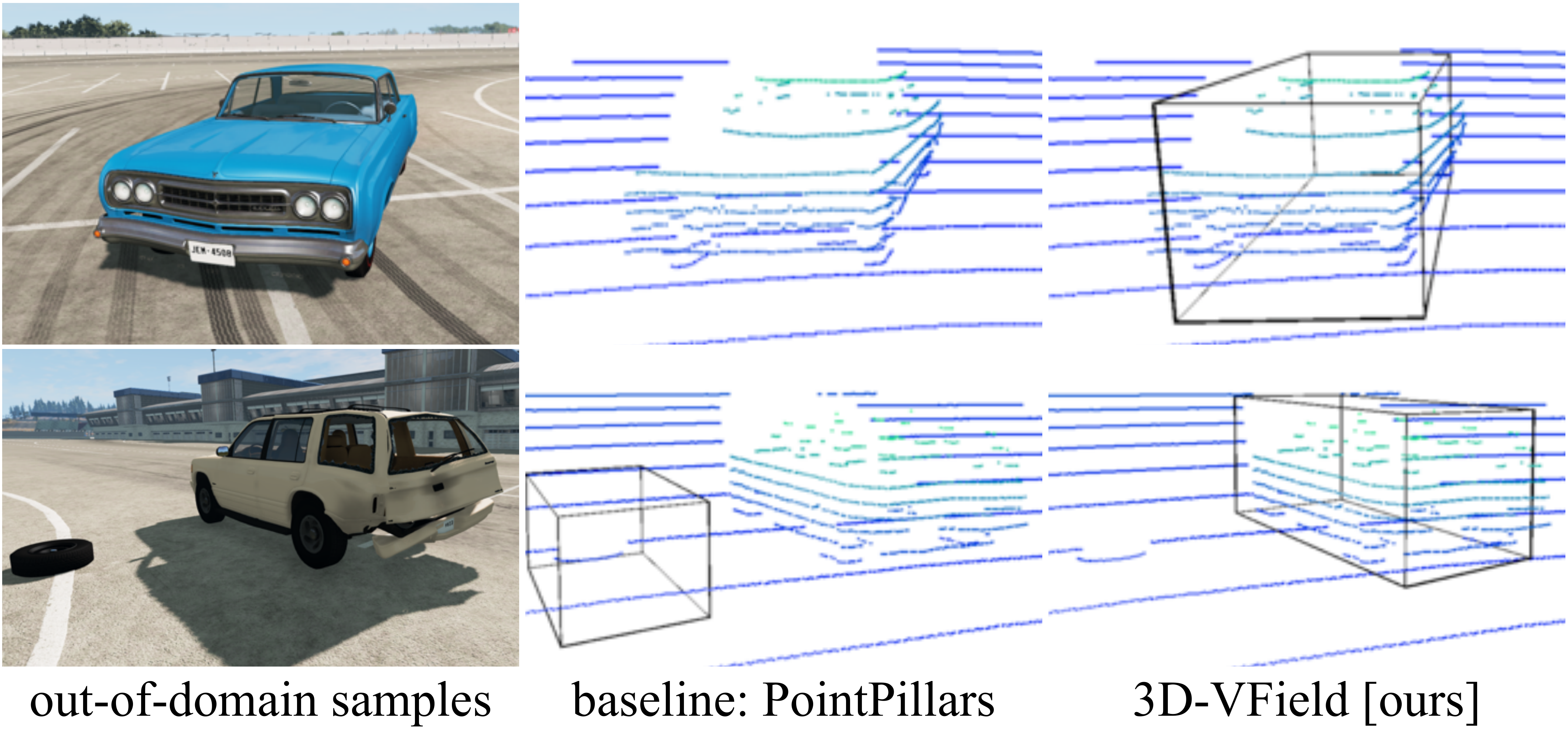}
\end{center}
   \caption{Predictions of PointPillars~\cite{lang_pointpillars_2019} trained on KITTI~\cite{geiger2012we}, without and with our adversarial augmentations on out-of-domain samples from the proposed CrashD dataset. CrashD comprises \textit{rare} (top) and \textit{damaged} (bottom) vehicles, resulting in natural adversarial examples~\cite{hendrycks2021natural}. As the models were applied to CrashD without fine-tuning, due to the different object shapes, the standard PointPillars delivered two false negatives and a false positive. \textit{Images used with courtesy of BeamNG GmbH.}}
\label{fig:teaser}
\end{figure}

Since corner cases are difficult to be captured as they occur in a dynamic real-life scenario, current datasets include only a limited amount of them, if any~\cite{bogdoll2021cornercases}, leaving most of these cases out-of-domain.
However, taking care of corner cases is particularly important in safety-critical settings, where long-tail and out-of-distribution samples could lead to dangerous issues if not accounted for during training~\cite{bogdoll2021cornercases}.

While several works have addressed some of these concerns on the imaging domain~\cite{qiao2020learning, beery2020synthetic, gasperini2021certainnet, hendrycks2021many}, this is still mostly unexplored for 3D point clouds~\cite{tu_physically_2020}, also due to the inherent challenges of point clouds, as they are unordered, sparse and irregularly sampled.
Nevertheless, as the output of 3D sensors (e.g., LiDAR, ToF cameras), point clouds are especially useful in high automation, where robustness and redundancy are intertwined with safety.

In this context, real non-standard objects, such as damaged and rare cars, or those from different regions, can lead to false negatives, as shown in Figure~\ref{fig:teaser}, since the inter-point geometry on which 3D detectors rely is different than usual.
While these examples can naturally occur in the real-world~\cite{hendrycks2021natural}, they can also be generated artificially with adversarial attacks~\cite{goodfellow_explaining_2015}.
This kind of approaches show the vulnerabilities of a model, which can then be addressed to improve robustness.
Recent adversarial point cloud alteration methods~\cite{tu_physically_2020} have tackled this problem to improve the generalization to out-of-distribution data.
However, despite being effective attacks, existing adversarial deformation strategies~\cite{xiang_generating_2019, liu2020adversarial} are sample-specific, lack wide-applicability, and by being designed without considering a 3D sensor, are mostly unconstrained in space~\cite{liu2020adversarial}.

In this work, we substantially improve the generalization capability of 3D object detectors to out-of-domain data, bridging this gap by deforming point clouds during training.
We propose 3D-VField: a novel adversarial augmentation method that learns to deform point clouds via widely-applicable and sample-independent vector fields (i.e., collections of vectors linked to a set of points in a given space).
Our deformations preserve the overall object shape, only slide points along the view ray, and do not add or remove any points. After learning a vector field, we use it to alter objects as data augmentation.
The main contributions of this paper can be summarized as follows:
\begin{itemize}
    \item We raise awareness on natural adversarial examples, such as those represented by damaged and rare cars, around their ability to fool popular 3D object detectors.
    \item We propose 3D-VField: a sensor-aware adversarial point cloud deformation method based on vector fields able to increase the generalization of 3D object detectors to out-of-domain samples via data augmentation.
    \item We introduce and publicly release CrashD: a dataset of damaged and rare cars. Extensive experiments on four outdoor and indoor datasets, namely KITTI~\cite{geiger2012we}, Waymo~\cite{sun2020waymo}, our CrashD, and SUN RGB-D~\cite{song2015sun}, show the wide applicability of our approach.
\end{itemize}

\section{Related Work}
\label{sec:related_work}
Our work is about adversarial augmentation to improve the generalization of 3D object detectors for point clouds.
In this section we provide a brief overview of existing approaches in these neighboring fields.

\subsection{Improving Generalization}
Generalization to unseen data is a highly desirable property for any learning-based approach~\cite{wang2021generalizing}.
Unseen data includes any samples on which a model has not been trained on, comprising both out-of-domain and in-domain data (e.g., validation set), depending on the size of the domain shift. In particular, domain generalization deals with improving the performance on a target domain, without any knowledge about it~\cite{wang2021generalizing}, in contrast to domain adaptation which has access to the target data~\cite{wang2020train}.
These works can be grouped in two broad categories: those acting on the model itself, and those operating on the input data.

Among the former category, model regularization strategies are commonly used to reduce overfitting~\cite{srivastava2014dropout} or address domain generalization~\cite{balaji2018metareg}. Estimating the model uncertainty was also found beneficial for out-of-domain data~\cite{gasperini2021certainnet}. Moreover, specific architectures can be found via search algorithms to improve robustness~\cite{mok2021advrush}.

A different category of works targets generalization by manipulating the input data. Towards this end, it is possible to leverage pretraining and multi-task learning to improve on out-of-distribution samples~\cite{albuquerque2020improving}. Additionally, synthetic data can be included to increase the accuracy on rare classes~\cite{beery2020synthetic}.
Data augmentation methods~\cite{summers2019improved, zhang2021mixup, hendrycks2021many} also belong to this category. Among these, there are adversarial approaches, which extended the training data with altered inputs learned in an adversarial fashion as a way to improve generalization~\cite{volpi2018generalizing, tu_physically_2020, qiao2020learning}.

The method we propose in this work addresses domain generalization (i.e., does not use any target information) and belongs to the data category, specifically to the adversarial approaches, which are detailed in Section~\ref{sec:rw_adversarial}.

\subsubsection{Generalization for 3D Object Detection}
In the context of generalization, some works addressed the task of 3D object detection, which is also the focus of this work.
Simonelli et al.~\cite{simonelli2020towards} created virtual views normalizing the objects with respect to their distance, to better generalize to samples at different depths in the image domain.
Tu et al.~\cite{tu_physically_2020} improved the generalization towards cars with roof-mounted objects, via adversarial examples on LiDAR point clouds.
Wang et al.~\cite{wang2020train} used domain adaptation to fill the gap between vehicles from multiple countries and different LiDAR sensors.


\subsection{Adversarial Examples}
\label{sec:rw_adversarial}
Adversarial examples are input alterations designed to lead a model to false predictions~\cite{szegedy_intriguing_2014, goodfellow_explaining_2015}.
A variety of works explored adversarial examples in the image domain~\cite{carlini_towards_2017, moosavi-dezfooli_deepfool_2016, papernot_limitations_2016, yuan_adversarial_2019, xiao_generating_2018}, where pixel perturbations imperceptible to humans are able to fool the target model.
Alaifari et al.~\cite{alaifari_adef_2019} deformed images using a different adversarial vector field learned for each sample. Wang et al.~\cite{wang_amora_2020} proposed adversarial morphing fields to alter image pixels spatially and fool classifiers.
However, this topic is still mostly unexplored on point clouds, especially those captured by 3D sensors (e.g., LiDAR, ToF camera).

\begin{figure}[t]
\begin{center}
\includegraphics[width=1.00\linewidth]{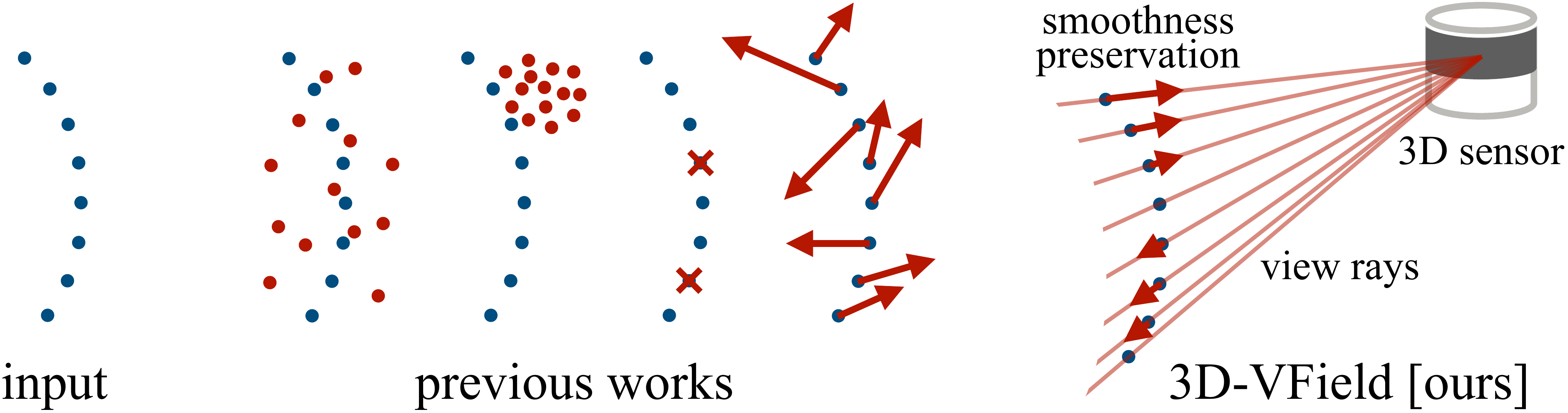}
\end{center}
   \caption{Adversarial deformations introduced by previous works, compared to ours. Other methods add, drop or move points with minor constraints. Ours only slides points along the view ray, while preserving shapes and occlusions.}
\label{fig:constraints}
\end{figure}

\begin{figure*}[t]
\begin{center}
\includegraphics[width=1.00\textwidth]{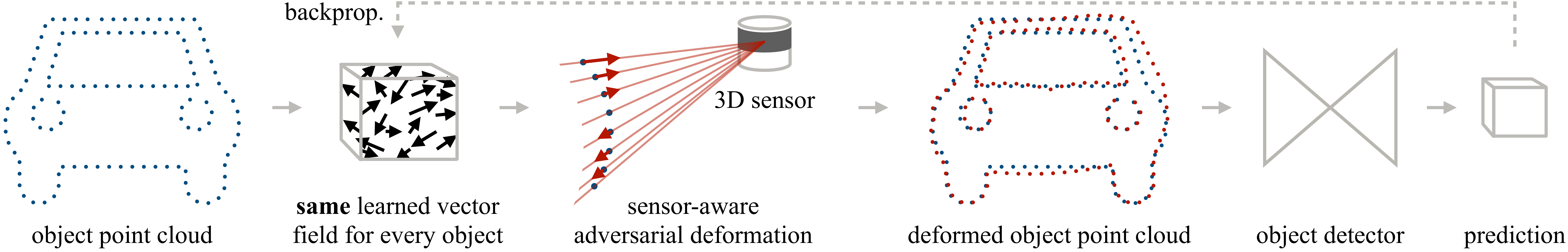}
\end{center}
   \caption{Overview of the proposed 3D-VField. We first learn a vector field adversarially to plausibly deform objects, taking constraints into account.
   The modified scenes are later used as augmentations to improve the generalization to unseen object shapes.
   }
\label{fig:attack_pipeline}
\end{figure*}

\subsubsection{Adversarial point clouds}
Adversarial methods for 3D point clouds can be grouped in three categories: generation if they add points, removal if they remove points, and perturbation if points are only shifted.
Then we present the methods from the perspective of generalization to out-of-domain samples.

\textbf{Generation and removal}
Xiang et al.~\cite{xiang_generating_2019} pioneered adversarial point clouds proposing a series of methods, some of which added points to fool the shape recognition.
Cao et al.~\cite{cao_adversarial_2019} showed the vulnerability of LiDAR-based methods against adversarial objects added to the scene.
Similarly, Tu et al.~\cite{tu_physically_2020} added adversarial meshes on top of cars.
A different line of works explored sensor attacks, adding points by means of a spoofing device~\cite{cao_adversarial_2019-1}. Conversely, removal methods adversarially learn to discard a few critical points~\cite{yang2019adversarial_removal}.

\textbf{Perturbation}
Xiang et al.~\cite{xiang_generating_2019} also proposed the first two adversarial perturbation approaches. One is the iterative gradient L2 attack, which is an adaptation of PGD from the image domain~\cite{madry2018pgd}, optimizing for a minimal deformation constrained by the L2 norm. Another approach is the Chamfer attack, which uses the Chamfer distance (CD) between the original and the deformed object to decrease the perceptibilty of the attack~\cite{liu2020adversarial}. The CD is measured by averaging the sum of the distances of the nearest neighbor from each point of the original point cloud to the deformed one. Using this distance function encourages point shifts across the surface of the object. Our method is closely related to the iterative gradient L2 attack, but we do not learn a vector for each point of each sample. Instead, we learn a sample-independent vector field and introduce further constraints to improve our deformations.
Liu et al.~\cite{liu2020adversarial} investigated perturbations more noticeable than the ones of Xiang et al., while producing continuous shapes by altering neighboring points accordingly.
Cao et al.~\cite{cao_invisible_2021} 3D printed adversarial objects to fool multi-modal (LiDAR and camera) detectors.

\textbf{Generalization}
Several works on adversarial point clouds were proposed targeting the ModelNet dataset~\cite{xiang_generating_2019, liu2020adversarial, hamdi2020advpc}, which comprises a set of synthetic 3D point clouds resembling various object shapes. Since ModelNet was not created with a 3D sensor, these foundation works often produce unrealistic outputs~\cite{xiang_generating_2019, liu2020adversarial}, that were not intended to improve the generalization of the models, but rather set the basis for adversarial attacks on point clouds~\cite{xiang_generating_2019}. Additionally, these mechanisms are sample-specific, making their applicability limited~\cite{xiang_generating_2019, liu2020adversarial, hamdi2020advpc}.
Instead, Tu et al.~\cite{tu_physically_2020} explored the impact on LiDAR object detection of meshed objects, such as canoes and couches, synthesized on top of a car roof. Moreover, they attacked these meshes in an adversarial fashion, and used them to defend the detector, thereby improving its robustness and generalization capability to unseen samples with roof-mounted objects.

Our work sets itself apart from all sample-specific methods~\cite{alaifari_adef_2019, xiang_generating_2019, liu2020adversarial, yang2019adversarial_removal}, as we construct a single highly transferrable and generic set of perturbations.
Similar to the work of Tu et al.~\cite{tu_physically_2020}, we aim to improve the generalization to out-of-domain samples. However, compared to theirs, as can be seen in Figure~\ref{fig:constraints}, we do not add any points, making ours a perturbation method.
Additionally, unlike Tu et al., by not making any assumptions on the object nor the kind of sensor, our method has a wider applicability, from indoor to outdoor settings. Plus, we improve realism by taking into account occlusion constraints, which were ignored so far, and making our deformations sensor-aware, as we only shift points along the sensor ray. Additionally, our method differs from all the ones above also because it generates adversarial point clouds via transferable learned vector fields, which has not been explored yet.

\section{Method}
We now illustrate our method, based on deforming point clouds to account for natural object variations, thereby improving the generalization of 3D object detectors to out-of-domain data via adversarial augmentation.
As shown in Figure~\ref{fig:attack_pipeline}, we achieve this by adversarially learning a vector field (Section~\ref{sec:adv_loss}).
Once trained, this vector field can be frozen and then applied to any previously seen or unseen objects, after scaling it to match the target size and constraining the points movement to preserve shapes and occlusions (Section~\ref{sec:field_application}). We apply it to deform all objects of its class, which we use as data augmentation (Section~\ref{sec:augmentation}).

\subsection{Adversarially learned vector field}\label{sec:vec_field}
We create a lattice of uniformly spaced 3D vectors within a 3D bounding box. Since the aim is to perturb the point cloud without adding or removing points, vectors are an immediate representation of this set of point shifts. This allows for both compactness and transferability, since the same learned vector field can be applied to any target object. To construct such a vector field, we discretize the space of a default bounding box $B_o$ with a step size $t$ to obtain root coordinates $\bm{f}$ in 3D space and assign an empty vector $\bm{v}= (x,y,z)$ to each root. $B_o$ is defined by width $w$, height $h$, length $l$, orientation angle $\alpha$ and its center $c = (x,y,z)$.

\textbf{Adversarial loss}\label{sec:adv_loss}
We use a binary cross entropy loss to suppress all relevant bounding box proposals, following~\cite{tu_physically_2020}. We consider a proposal as relevant if the prediction confidence score $s > 0.1$. $\mathcal{Q}$ is the set of relevant proposal $q$, where each $q$ has a confidence score $s$. We minimize $s$, weighed by the 3D IoU with the the ground truth $q^*$:
\begin{align}
\mathcal{L}_{\rm adv} = \sum_{\substack{q, s~\in~\mathcal{Q}}} - \mathrm{IoU}(q^*, q) \log(1 - s).
\end{align}
By repeatedly reducing the confidence score while training the vector field, the detector misses the object or predicts a misaligned box.
During training, we apply the same vector field to each target object in every scene, minimizing the loss on the whole dataset.
At each optimization step, the vectors are updated, resulting in differently deformed point clouds of target objects, which eventually lead to different predictions. As $\mathcal{L}_{\rm adv}$ smoothly converges, the performance of the detector, against which the vector field is optimized, decreases. Once trained, the vectors can be used for data augmentation.

\subsection{Objects Deformation}
\label{sec:field_application}
Before applying a vector field, we scale it to match the target object size. Manipulating the points through these vectors, we constrain their movement as described below.

\textbf{Optical ray consistency}
To help generalization and preserve the sensor's physical constraints when generating deformations, we employ a simple sensor model in which the 3D points can only be moved across the optical ray. We first compute the ray $\bm{u}_i$ between the 3D sensor and each point $\bm{p}_i$, which determines the deformation direction for each point.
Then we calculate the deformation vectors $\bm{r}_{i}$, for each $\bm{p}_i$ by projecting its nearest vector $\bm{v}_{i}$ onto the ray $\bm{u}_i$. Points are therefore only moved by $\bm{r}_{i}$.

\textbf{Regularizing the deformations}\label{sec:constraints}
We limit the perturbation of the points by restricting the vectors with $\Vert \bm{v} \Vert_{\infty} < \epsilon$ following the standard PGD $L_\infty$ attack~\cite{madry2018pgd}.
We then ensure shape smoothness along the object surface by sampling multiple $k$ neighboring vectors to move a given 3D point. For each $j$-th nearest neighbor we calculate the euclidean distance $d_{ij}$ between each point $\bm{p}_i$ of the object and its nearest vector $\bm{v}_{ij}$ from the vector field.
The final shift $\bm{m}_i$ of each point is calculated by weighting the deformation vectors $\bm{r}_{ij}$ with their corresponding distance $d_{ij}$:
\begin{align}
\bm{m}_i = \frac{\sum_{j=1}^{k} d_{ij} \bm{r}_{ij} }{k}
\end{align}
This allows for a more gradual depth difference between neighboring points, as neighboring vectors with opposite directions would lead to almost no movement of the affected point. Thus, shape smoothness is preserved and less irregular deformations are produced.

\textbf{Relative rotation}
We found that using a single vector field for all objects present in the dataset leads to very low amounts of deformation. Due to the various object poses, its vectors would be pointing in all directions, decreasing its efficacy.
We circumvent this and allow for a larger degree of alignment between neighboring vectors, by first clustering all the objects in the dataset w.r.t. the relative orientation between object and sensor, and then learning $G$ different fields, one for each cluster.

\subsection{Adversarial Data Augmentation}
\label{sec:augmentation}
During training of the object detector, we perturb the input point clouds by using the adversarially learned vector fields as data augmentation. This increases the robustness, given that the learned deformations are structurally-consistent, and are therefore more capable than standard augmentations (e.g., scaling, flip, rotation) of resembling out-of-domain car shapes, such as vehicles from a different country~\cite{wang2020train}.
We increase the variability by learning $N$ different vector fields for each of the $G$ rotations (Section~\ref{sec:field_application}).
During training, we randomly select only one object in the scene, and we deform it with a randomly chosen vector field out of the $N$ possible ones for its relative rotation.
This high variability ensures that the model learns both normal and deformed objects, and that each sample can be deformed differently across training, thereby preventing overfitting to specific deformations.

\section{Experiments and Results}

\subsection{Experimental Setup}

\textbf{Datasets}
We conducted our experiments on four different datasets. Three of them are autonomous driving LiDAR-based: KITTI~\cite{geiger2012we}, the Waymo Open Dataset~\cite{sun2020waymo}, and the proposed synthetic CrashD, which we introduce below. Additionaly, we apply our method also on the indoor SUN RGB-D dataset~\cite{song2015sun}, showing its wide applicability.
\textbf{KITTI} is a popular 3D object detection benchmark recorded in Germany. We adopted a standard split~\cite{lang_pointpillars_2019}, which comprises 3712 training and 3769 validation LiDAR point clouds, where we used the \textit{car} class, reporting on the standard \textit{easy}, \textit{moderate} and \textit{hard}.
We evaluated models trained on KITTI (without any fine-tuning) on Waymo and our CrashD to assess the generalization capability of the models to out-of-domain data, particularly critical for autonomous driving.
The \textbf{Waymo} dataset is a challenging large-scale collection of real scenes recorded in various locations of the USA. It is highly diverse with different weather and illumination conditions, such as rain and night.
Furthermore, in the Supplementary Material we show the wide-applicability of our techniques on time-of-flight (ToF) cameras with the \textbf{SUN RGB-D} dataset.

\textbf{CrashD dataset}
To quantify the generalizability on out-of-domain samples, we produced a synthetic dataset named CrashD. As this includes various types of cars, such as normal, old, sports and damaged, it comprises a variety of plausible vehicle shapes, thereby serving as a valuable out-of-domain test.
Specifically, the crashes are individually generated with a realistic simulator~\cite{BeamNGTechnicalPaper21} and distinguished depending on the intensity, namely \textit{light}, \textit{moderate}, \textit{hard}, as well as the kind of damage: \textit{clean} (i.e., undamaged), \textit{linear} (i.e., frontal or rear), and \textit{t-bone} (i.e., lateral). The randomly and automatically generated 15340 scenes were captured by a 64-beam LiDAR configured to mimic the KITTI one. Each scene presents between 1 and 5 vehicles, with visible damages, before being repaired and placed at the same locations to collect the \textit{clean} set, resulting in a total of 46936 cars.
We are releasing this data publicly, as an out-of-domain evaluation benchmark for models trained on KITTI~\cite{geiger2012we}, Waymo~\cite{sun2020waymo} or similar datasets.
Further details can be found in the Supplementary Material.


\textbf{Evaluation metrics}
We evaluated the object detection performance on the standard \textbf{AP}, with a 3D IoU threshold of 0.7 for KITTI and CrashD, 0.5 for Waymo, and the standard 0.25 for SUN RGB-D. To measure the quality of the adversarial perturbations we followed Tu et al.~\cite{tu_physically_2020} using the \textbf{attack success rate} (ASR) metric. It measures the percentage of objects that become false negatives after undergoing an adversarial alteration.
For the ASR, we considered an object detected if its 3D IoU was larger than 0.7.

\begin{table*}[t]
\begin{center}
\begin{tabular}{l|l|ccc|r|c|cc|cc}
\toprule

\multicolumn{2}{c|}{~} & \multicolumn{4}{c|}{KITTI} & $\rightarrow$ Waymo & \multicolumn{4}{c}{$\rightarrow$ CrashD} \\

\multicolumn{2}{c|}{~} & \multicolumn{3}{c|}{AP} &  & & \multicolumn{2}{c|}{AP \textit{normal}} & \multicolumn{2}{c}{AP \textit{rare}} \\
\multicolumn{1}{l|}{Architecture} & \multicolumn{1}{l|}{Method} & \textit{easy} & \textit{mod.} & \textit{hard} & ASR & AP & \textit{clean} & \textit{crash} & \textit{clean} & \textit{crash} \\

\midrule

\multirow{10.4}{*}{PointPill.~\cite{lang_pointpillars_2019}} 
& no augm.~\cite{lang_pointpillars_2019} & 70.00 & 61.88 & 56.23 & -~~~ & 30.68 & \multicolumn{1}{r}{1.79} & \multicolumn{1}{r|}{0.93} & \multicolumn{1}{r}{3.92} & \multicolumn{1}{r}{2.33} \\
& no obj.~sampl.~\cite{lang_pointpillars_2019} & 83.83 & 74.14 & 68.30 & -~~~ & 37.85 & 50.36 & 36.44 & 28.70 & 20.02 \\
& PointPillars~\cite{lang_pointpillars_2019} &  \textbf{88.24} & 77.11 & 74.55 & -~~~ & 40.86 & 65.20 & 43.67 & 34.14 & 22.48  \\

& iter.~grad.~L2~\cite{xiang_generating_2019} &  86.24 &76.92 & 73.84 & $^*$95.9 & 39.86 & 58.65 & 41.86 & 35.92 & 23.69 \\
& Chamfer att.~\cite{liu2020adversarial} &  87.15 & 77.05 & 74.07 & $^*$\textbf{99.8} & 40.54 & 56.84 & 39.56 & 36.29 & 24.73\\
& advers.~gener.~\cite{xiang_generating_2019} & 86.12 & 76.39 & 73.18 & $^*$91.6 & 40.55 & 57.75 & 38.03 & 35.73 & 24.18 \\
& advers.~remov.~\cite{yang2019adversarial_removal} & 86.51 & 76.85 & 74.04 & $^*$86.1 & 40.32 & 66.52 & 48.88 & 41.42 & 28.10 \\
& 3D-VField [ours] & 87.05 & \textbf{77.13} & \textbf{75.55} & 63.4 & \textbf{44.61} & \textbf{67.95} & \textbf{52.87} & \textbf{43.40} & \textbf{30.37} \\

\cmidrule{2-11}
& SN dom.~adapt.~\cite{wang2020train} & -&-&-&-~~~ & 49.27 & 79.42 & 72.59 & 60.53 & 48.23 \\
& [ours] + SN~\cite{wang2020train} & -&-&-&-~~~ & \textbf{51.32} & \textbf{92.14} & \textbf{87.28} & \textbf{86.26} & \textbf{76.42} \\

\midrule
\multirow{2}{*}{Second~\cite{yan_second_2018}} & Second~\cite{yan_second_2018} &  \textbf{88.93} & \textbf{78.68} & \textbf{76.87} & -~~~ & 42.45 &  72.73 & 56.74 & 41.85 & 32.84\\
& 3D-VField [ours]  & 88.87 & 78.56 & 76.81 & 54.9 & \textbf{43.51}  & \textbf{76.54} & \textbf{60.51} & \textbf{47.47} & \textbf{36.14}\\

\midrule
\multirow{2}{*}{Part-A$^2$~\cite{shi2019parta2}} & Part-A$^2$~\cite{shi2019parta2} & 89.60 & 79.16 & 78.52 & -~~~ & 49.76 & 83.05 & 63.25 & 74.03 & 52.33\\
& 3D-VField [ours] & \textbf{89.65} & \textbf{79.26} & \textbf{78.62} & 50.5 & \textbf{56.08}  & \textbf{88.80} & \textbf{73.80} & \textbf{81.10} & \textbf{61.34}\\

\bottomrule

\end{tabular}
\end{center}
\caption{Comparison of models trained on KITTI~\cite{geiger2012we} towards out-of-domain data (without any fine-tuning), namely Waymo validation set~\cite{sun2020waymo} and our CrashD datasets, as well as on the KITTI validation set. Each method applies a data augmentation (for adversarial ones ASR is measured on their adversarial examples), or performs domain adaptation (only SN~\cite{wang2020train} in this work), resulting in the reported APs.
$\rightarrow$: transfer from KITTI. $^*$: being sample-specific, the adversarial method had to be trained on the validation set of KITTI.
}
\label{table:AP_on_all}
\end{table*}

\textbf{Network architectures}
We used four different 3D object detectors.
PointPillars~\cite{lang_pointpillars_2019} voxelizes the scene in vertical columns (i.e., pillars) from the bird's eye view, using PointNet for feature extraction.
Second~\cite{yan_second_2018} voxelizes the point cloud and uses a learned voxel feature encoding. Part-A$^2$ Net~\cite{shi2019parta2} is an extension of PointRCNN that predicts intra-object part locations for improved accuracy. VoteNet~\cite{qi2019_votenet} (Supplementary Material) is based on PointNet++ and Hough voting. While the first three are mostly used for autonomous driving, VoteNet is used indoor.

\textbf{Implementation details}
We constructed each vector field within $B_o$ with $w=1.8$m, $h=1.6$m, $l=4.6$m and a step size of $t = 20$cm resulting in 1656 vectors per vector field. If not stated otherwise, we grouped objects by relative rotations with $G = 12$ groups, and set $N = 6$.
During the perturbation stage, we moved points according to their $k=2$ nearest vectors and deform only along the sensor ray. 
For the PGD optimization, we used Adam with a learning rate of 0.05. The distance threshold was set to $\epsilon = 30$cm. Each vector was randomly initialized form a uniform distribution with values between -1cm and 1cm.
We trained all models using PyTorch and MMDetection3D~\cite{mmdet3d2020} on a single NVIDIA Tesla V100 32GB GPU.

\textbf{Prior works and baseline}
We focused on object detection and compared with other adversarial methods.
All models were applied on PointPillars~\cite{lang_pointpillars_2019}, unless otherwise noted.
As point perturbation methods we used the iterative gradient L2~\cite{xiang_generating_2019} and the Chamfer attack~\cite{liu2020adversarial}. For generation we used \cite{xiang_generating_2019} adding 10\% and \cite{yang2019adversarial_removal} removing 10\% of the objects points.
For a fair comparison, we trained all on the same KITTI dataset split~\cite{lang_pointpillars_2019},
with $\epsilon = 30$cm, then we altered the point clouds as data augmentation with the same settings as ours (i.e., random selection of one object per scene to augment). Moreover, we combined ours with the domain adaptation statistical normalization (SN) strategy of~\cite{wang2020train}. Following~\cite{wang2020train}, after computing the average box dimensions in the target datasets (i.e., Waymo and CrashD), we scaled the source (i.e., KITTI) point clouds within the ground truth boxes accordingly and fine-tuned the trained models with this altered target-aware source data.

\subsection{Quantitative Results}
\label{sec:quantitative}
\textbf{Adversarial methods and generalization}
Table~\ref{table:AP_on_all} shows the comparison between our 3D-VField and related adversarial approaches when applied on PointPillars~\cite{lang_pointpillars_2019} in the context of generalization. In particular, we report other adversarial perturbation methods, such as the iterative gradient L2~\cite{xiang_generating_2019} and the Chamfer attack~\cite{liu2020adversarial}, adversarial generation~\cite{xiang_generating_2019}, as well as adversarial removal~\cite{yang2019adversarial_removal}.
Augmenting with the adversarial examples of our 3D-VField did not reduce the overall in-domain AP compared to PointPillars, but brought numerous benefits in terms of out-of-domain generalization.
As demonstrated by Wang et al.~\cite{wang2020train}, the transfer from KITTI to \textbf{Waymo} is particularly challenging due to the different shapes and sizes of the vehicles found in Germany and the USA, as well as the 50\% higher point density and the narrower field of view~\cite{sun2020waymo}. This test assesses the quality of the generated deformations with respect to real vehicle shapes found in a different country. On Waymo our 3D-VField delivered more than 9\% relative improvement over PointPillars and the other adversarial methods, and 13\% over Part-A$^2$~\cite{shi2019parta2}, proving the benefit of our added sensor-awareness on real and challenging out-of-domain data.
On the right of Table~\ref{table:AP_on_all} we report the results on the proposed \textbf{CrashD}. It can be seen that despite the transfer from KITTI, the AP on \textit{clean normal} cars is relatively high for all approaches, likely because those samples are not particularly difficult. However, when damaging those exact same vehicles and placing them at the same locations (\textit{crash}), the detection performance dropped. This shows the effort required for the methods to relate these to the cars learned on KITTI, and proves them as natural adversarial examples. Similarly, with \textit{rare} cars (i.e., old and sports cars), the AP dropped even more, quantifying the domain shift from \textit{normal} vehicles. \textit{Rare crash} cars, by combining the two out-of-domain aspects (i.e., rarity and damage), were the hardest for all methods, reducing the AP from \textit{normal clean} by up to two thirds (PointPillars). Nevertheless, our method improved significantly over the detectors and the other adversarial approaches for all transfers and categories. This can be attributed to our adversarial augmentations introducing diversity in the training data, while being sensor-aware. In particular, the sensor-awareness ensures that the deformed point clouds are still plausible, thereby better resembling possible out-of-domain samples, such as those of Waymo and CrashD. Among the other adversarial approaches, only removing points~\cite{yang2019adversarial_removal} improved generalization to CrashD, probably because it preserved the overall point clouds. Nevertheless, \cite{yang2019adversarial_removal} was not beneficial on Waymo, which features denser point clouds and more challenging real scenes.

\textbf{Combination with data augmentations}
As adversarial data augmentation, our 3D-VField is not alternative to different augmentation strategies, but can be applied in combinations with others.
In Table~\ref{table:AP_on_all} we show how common data augmentation techniques impact the detections for PointPillars~\cite{lang_pointpillars_2019}. Using no augmentations (no augm.) critically reduced the APs, especially on CrashD at IoU 0.7 (Table~\ref{table:AP_on_all}). At IoU 0.5, this resulted in an AP on \textit{normal clean} of 65.59, while the baseline~\cite{lang_pointpillars_2019} delivered 98.91. Introducing standard augmentations (no obj.~sampl., e.g., flip and rotation) improved, but adding the popular object sampling~\cite{lang_pointpillars_2019} (PointPillars) increased the APs further. On top, our augmentations substantially improved all transfers, without decreasing the in-domain performance.

\textbf{Combination with domain adaptation}
By addressing domain generalization, our approach does not use any target information. Therefore, ours is not alternative to domain adaptation methods~\cite{wang2020train}, which make use of target data. However, similarly to other data augmentation strategies, our 3D-VField can be combined with domain adaptation techniques. As shown in Table~\ref{table:AP_on_all}, such combination further boosts the performance on challenging out-of-domain data. By altering the objects size via the statistical normalization (SN) of~\cite{wang2020train}, the AP on Waymo increased. Constrained by the high amount of false positives and negatives, when combined with SN, ours retained a margin of over 2\% compared to PointPillars with SN.
Moreover, the AP on CrashD improved dramatically across all categories, especially for the hardest \textit{rare crash} group. The results show how, despite a substantial increase in AP from PointPillars~\cite{lang_pointpillars_2019}, SN alone did not reach the full potential of the detector. Only when combined with ours, the AP doubled (\textit{normal crash}) and more than tripled (\textit{rare crash}) over PointPillars, without using any extra target information. This shows the benefit of this combination, and reiterates the added value of incorporating adversarially deformed objects via data augmentation to improve generalization to out-of-domain samples.

\begin{table}[t]
\begin{center}
\begin{tabular}{l|c|cc|r}
\toprule

\# $G$ & K. ASR $\uparrow$ & K. \textit{mod.} & $\rightarrow$ Waymo & \# vectors \\

\midrule

1 & 55.08 & \textbf{77.32} & 40.43 & \textbf{10K} \\ 
12 & \textbf{63.37} & 77.13 & \textbf{44.61} & 120K \\
360 & 44.84 & 77.06 & 40.30 & 3.6M \\
\bottomrule
\end{tabular}
\end{center}
\caption{
Our 3D-VField trained on KITTI (K.) with varying amounts of relative rotations $G$. $\rightarrow$: transfer no fine-tuning.
}
\label{table:relative_rotations}
\end{table}

\textbf{Adversarial methods as attacks}
In terms of ASR (Table~\ref{table:AP_on_all}), our approach is not as strong as the other adversarial methods, namely the iterative gradient L2~\cite{xiang_generating_2019}, the Chamfer attack~\cite{liu2020adversarial}, adversarial generation~\cite{xiang_generating_2019} and removal~\cite{yang2019adversarial_removal}. However, this is expected as our vector fields are sample-independent, compared to their point-to-point deformations being sample-specific. Due to this reason, their alterations had to be learned directly on the KITTI validation set, on which the ASR was measured. Nevertheless, a very high ASR means the altered objects are unrecognizable, which does not aid generalization. The goal of our method is not having a detector fully miss the attacked objects (high ASR), but rather deforming them to improve the performance on out-of-domain data. Towards this end, the perturbed objects need to be at the same time altered enough to add diversity to the training data, and not be too far apart from the training distribution to avoid confusing the detector. We found this balance by learning our vector fields adversarially, while preserving the objects shape and the sensor realism with our added constraints.

\begin{table}[t]
\begin{center}
\begin{tabular}{l|r|r|r|cc}
\toprule
 & \multicolumn{2}{c|}{KITTI} & \multicolumn{1}{c|}{$\rightarrow$ W.} & \multicolumn{2}{c}{$\rightarrow$ CrashD} \\

Method & \multicolumn{1}{c|}{\textit{mod.}} & ASR & & \textit{n.,clean} & \textit{r.,crash} \\

\midrule

P.P.~\cite{lang_pointpillars_2019} & 77.11 & -~~~ & 40.86 & 65.20 & 22.48  \\

no learn & 76.36 & 10.1 & 41.62 & 62.94 & 21.75 \\

unleash & 76.82 & \textbf{97.7} & 40.95 & 60.43 & 27.55 \\

ray con.  & 76.35 & 59.5 & 41.03 & 59.82 & 29.16 \\

full & \textbf{77.13} & 63.4 & \textbf{44.61} & \textbf{67.95} & \textbf{30.37} \\


\bottomrule
\end{tabular}
\end{center}
\caption{Ablation on the deformation constraints imposed by our method, compared to PointPillars (P.P.)~\cite{lang_pointpillars_2019}.
Trained on KITTI. $\rightarrow$: transfer no fine-tun.; W.: Waymo.
}
\label{table:ablation_learning}
\end{table}

\textbf{Different 3D detectors}
In Table~\ref{table:AP_on_all}, we also compare the performance of our 3D-VField when paired with different 3D object detectors, namely PointPillars~\cite{lang_pointpillars_2019}, Second~\cite{yan_second_2018}, and Part-A$^2$~\cite{shi2019parta2}. Remarkably, using the proposed adversarial augmentation improved the AP of Part-A$^2$ on Waymo by a large margin. The superiority of Part-A$^2$ over the other detectors can be attributed to its part-awareness~\cite{shi2019parta2}, which might have set its focus on the most relevant object parts (e.g., wheels) and their relationships to identify cars also in out-of-domain settings. For Second~\cite{yan_second_2018}, the performance on KITTI turned out lower than the one reported in~\cite{mmdet3d2020}, despite using the same settings and framework. This reduced AP affected both the baseline~\cite{yan_second_2018} and our approach. Nevertheless, adding our adversarial deformations significantly improved the generalization of all three detectors to out-of-domain data, despite training our vector fields solely against PointPillars. This shows the wide applicability and transferability of our techniques.

\begin{figure*}[t]
\centering
  \includegraphics[width=1.0\textwidth]{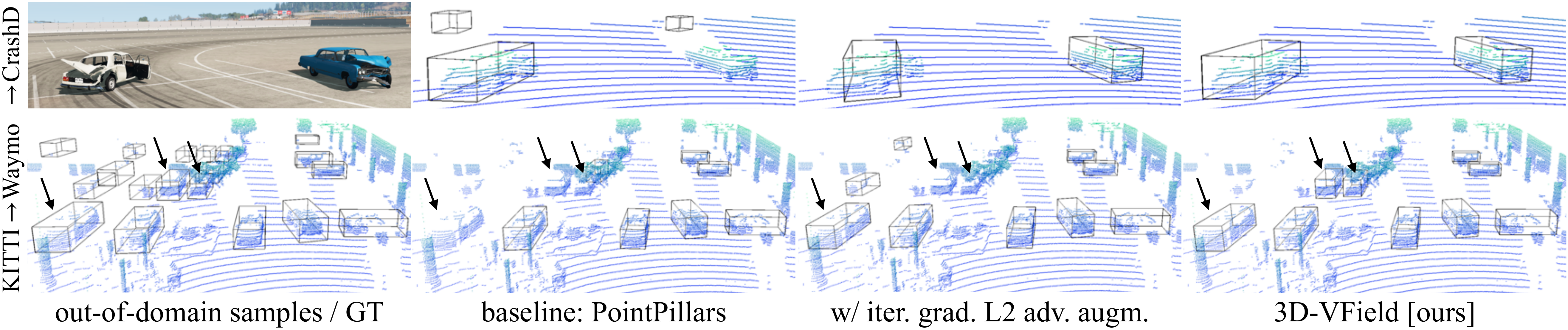}
   \caption{
   Predictions on challenging out-of-domain samples from the proposed CrashD (top) and Waymo~\cite{sun2020waymo} (bottom). Models based on PointPillars~\cite{lang_pointpillars_2019} trained on KITTI (without fine-tuning). Iterative gradient L2~\cite{xiang_generating_2019} and ours trained with adversarial augmentation.
   }
   \label{fig:qualitative}
\end{figure*}

\textbf{Specificity-generalization trade-off}
Table~\ref{table:relative_rotations} shows that by varying the amount of relative rotations $G$, a trade-off arises between generalization, attack specificity (i.e., strength on individual samples by overfitting to the training data), and storage (i.e., amount of vectors). $G=12$ offers a good balance. With the extreme $G=$\# of objects, ours would become sample-specific, inheriting the weaker generalization capabilities of~\cite{liu2020adversarial,yang2019adversarial_removal}. While these methods needed to be trained on the validation set, allowing for high ASRs (Table~\ref{table:AP_on_all}), our vectors were learned on the training set. So with high $G$, ours overfitted on the training data, which is visible evaluating on the validation set.
Our augmentation strategy learns only 1656 3D vectors to perturb objects. However, by training with $G=12$ and $N=6$, the amount of vectors increased to 120K. Conversely, the sample-specific iterative gradient L2~\cite{xiang_generating_2019} and the Chamfer~\cite{liu2020adversarial} attacks required 10.9M and 12.6M vectors for training and validation sets respectively. This shows the easy applicability of our 3D-VField.

\textbf{Ablation study on deformation constraints}
As we introduced the sensor-awareness and the surface smoothness constraints to our deformations, we investigate their impact in terms of generalization to out-of-domain data. In Table~\ref{table:ablation_learning}, we report this comparison when limiting the deformations to $\epsilon = 30$ cm.
It can be seen that not learning the perturbations, but applying all our constraints (no learn) could already be a beneficial augmentation technique, as it improved the transfer to Waymo.
Instead, removing all constraints, but learning the vector fields (unleash) delivered a strong ASR of 97.7\%. This significantly increased the AP on the CrashD \textit{rare} cars.
When deforming with sensor-awareness (ray con.), ASR reduced, but the AP on the most difficult transfer settings (i.e., \textit{rare crash}) improved.
Our full model 3D-VField, adds the distance smoothing (Section~\ref{sec:constraints}) delivering superior transfer capabilities.
Furthermore, increasing the maximum deformation $\epsilon$ to 40 or 60 cm, improved the ASR to 73.3\% and 87.1\%, but as augmentation decreased the AP on KITTI by 1\% and 1.7\%, respectively. This means that higher deformations do not generalize well, as their plausibility decreases, while 30 cm offers a good trade-off.

\subsection{Qualitative Results}
In Figure~\ref{fig:qualitative} we compare the transfer predictions from KITTI to CrashD and Waymo~\cite{sun2020waymo} of the standard PointPillars~\cite{lang_pointpillars_2019}, augmented with ours and the iterative gradient L2 adversarial approach~\cite{xiang_generating_2019},  which is the closest to ours in terms of adversarial deformation (Section~\ref{sec:related_work}). For CrashD, as seen in the quantitative results (Section~\ref{sec:quantitative}), the iterative gradient L2 method delivered better detections compared to not using any adversarial augmentations~\cite{lang_pointpillars_2019}, but our 3D-VField outperformed it, with a more aligned box for the left damaged car. The figure also shows the severity of the \textit{hard} damages present in CrashD, and how adversarial augmentation helps to detect such challenging samples.
For the difficult transfer KITTI $\rightarrow$ Waymo (Section~\ref{sec:quantitative}), it can be seen that all methods had troubles detecting the cars with few points in the parking lot on the left. Furthermore, PointPillars~\cite{lang_pointpillars_2019} ignored 3 recognizable cars with a high amount of points, while augmenting with the iterative gradient L2 caused missing 2 of them and detecting 2 further ones, albeit with misaligned boxes. Instead, despite missing further ones, our method was able to recognize these visible cars.

\begin{figure}[h]
    \begin{center}
        \includegraphics[width=1.\linewidth]{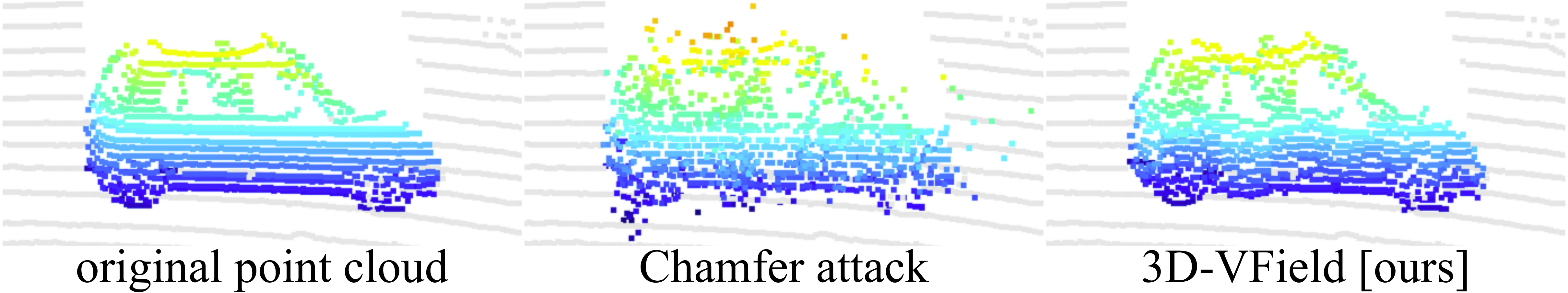}
    \end{center}
    \caption{Example deformations by our method and the Chamfer attack~\cite{liu2020adversarial} on a car of the KITTI validation set~\cite{geiger2012we}.}
    \label{fig:deforamations_comp_paper}
\end{figure}

Figure~\ref{fig:deforamations_comp_paper} confirms that the strong ASR of the Chamfer attack~\cite{liu2020adversarial} seen in Table~\ref{table:AP_on_all} corresponds to unrecognizable objects. It also provides an example of the minor deformations introduced by our adversarial vector fields. By preserving the overall shape of the car and its surfaces, ours allowed for superior generalization to unseen data.

We refer to the \textbf{Supplementary Material} for more results on indoor  settings, transferability, robustness against noise, detailed evalutations on CrashD, and various ablation studies on grouping and aggregation strategies, as well as the amount of deformed objects during training.

\section{Conclusion}
In this paper we presented 3D-VField: an adversarial augmentation method for point clouds to improve the object detection performance on natural adversarial examples and out-of-domain data, such as rare, damaged cars, or vehicles from different regions.
Towards this end, 3D-VField produces plausible shapes used as data augmentation.
Extensive experiments showed the high generalization and transferability of the proposed approach, from indoor to outdoor settings, on both real and synthetic data.
Furthermore, we proposed and released CrashD: a new benchmark to challenge 3D object detectors on out-of-domain data, including various kinds of damaged and rare cars.

\appendix
\section{Supplementary Material}
In this supplementary material we include further details and results. Specifically, Section~\ref{sec:crashd_full_blown} describes the proposed CrashD out-of-domain dataset to a greater extent, Section~\ref{sec:appendix_impl} provides additional implementation details, Sections~\ref{sec:appendix_quantitative} and~\ref{sec:appendix_qualitative} report more quantitative and qualitative results on outdoor data, while Section~\ref{sec:indoor_appendix} provides results on ToF camera data in indoor settings.

\subsection{Details on the Proposed Dataset: CrashD}\label{sec:crashd_full_blown}
In this section we further describe the proposed dataset: CrashD.
We refer the reader to the dataset webpage to see examples of the generated accidents and scenes.

\subsubsection{Intended Use}
This dataset was designed to evaluate the performance of LiDAR-based 3D object detectors on out-of-domain data.
It is meant to serve as a test benchmark for 3D detectors trained on KITTI~\cite{geiger2012we}, Waymo~\cite{sun2020waymo}, or similar datasets.

It should be noted, that CrashD is not intended for training and evaluating an object detector directly, since the generated LiDAR scenes do not include anything other than ground and cars. Therefore, training and evaluating on this dataset would be rather trivial, since the detector could learn that anything rising from the ground is a car, except for the relatively small spare parts separated by the accidents (e.g., the tire in Figure~\ref{fig:teaser}). 

Nevertheless, reasonable uses of the proposed CrashD could include domain adaptation, transfer learning, and domain generalization~\cite{wang2021generalizing}, as well as synthetic-to-real transfers. Furthermore, it could be used to assess the damage of a vehicle, and also for uncertainty estimation or similar methods to detect out-of-distribution samples. Moreover, it could serve for point cloud reconstruction, or anomaly segmentation approaches comparing damaged and undamaged cars, since for each crashed vehicle in a scene we provide its repaired counterpart at the same location.

\subsubsection{Driving Simulator}
CrashD was generated using a driving simulator developed by BeamNG~\cite{BeamNGTechnicalPaper21}, which includes a realistic physics engine, allowing for realistic damages. It offers a Python interface to setup the scenarios programmatically. Furthermore, it features a variety of sensors, including a LiDAR with customizable settings. Therefore, we equipped the ego vehicle with a LiDAR that imitates the one used in KITTI~\cite{geiger2012we}.

\begin{figure}[t]
\begin{center}
\includegraphics[width=1.00\linewidth]{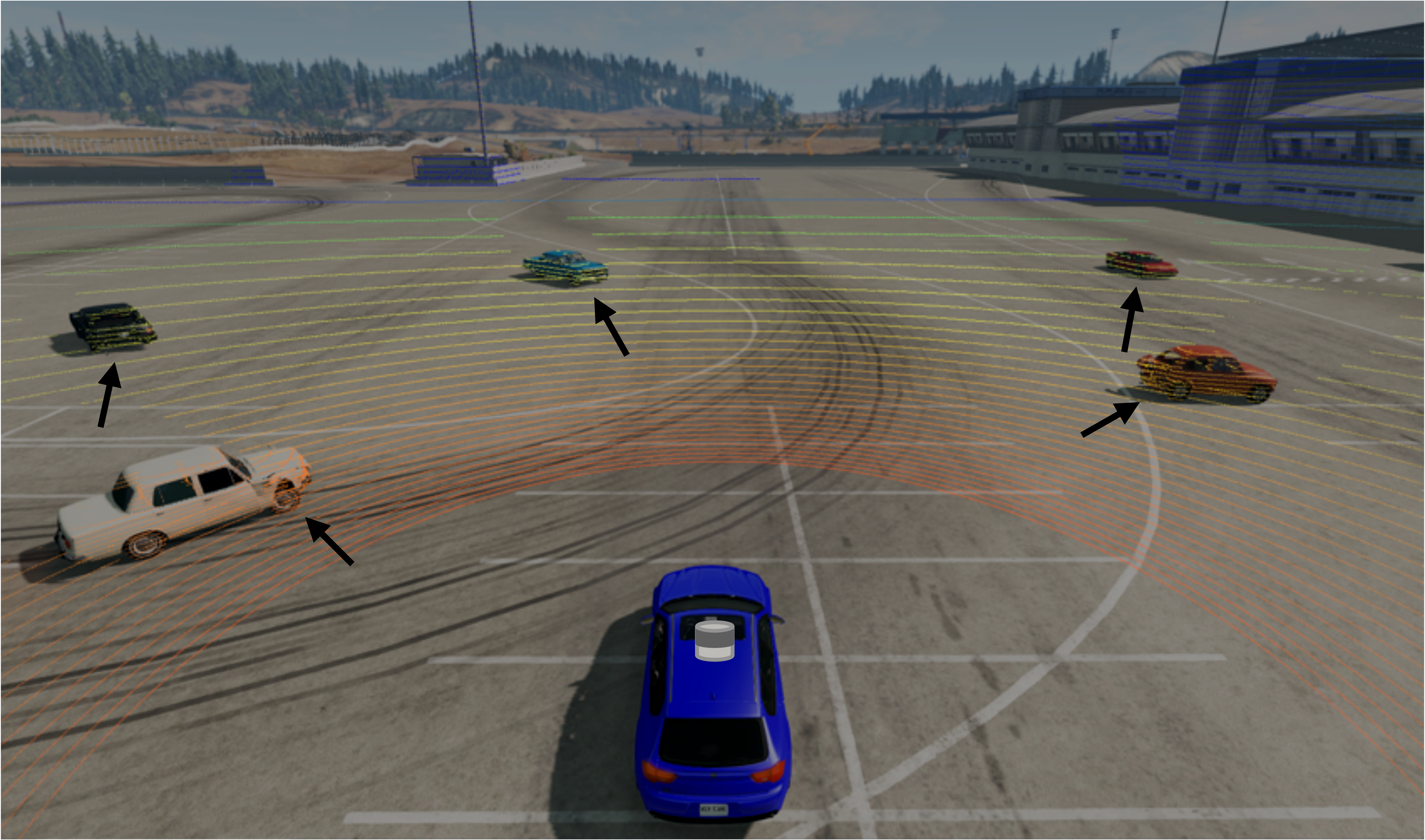}
\end{center}
   \caption{LiDAR scene setup of CrashD. For each car, a black arrow indicates its damaged area, which is ensured to be visible from the sensor viewpoint. \textit{Image used with courtesy of BeamNG GmbH.}}
\label{fig:lidar_scene}
\end{figure}

\begin{figure*}[t]
\centering
  \includegraphics[width=1.0\textwidth]{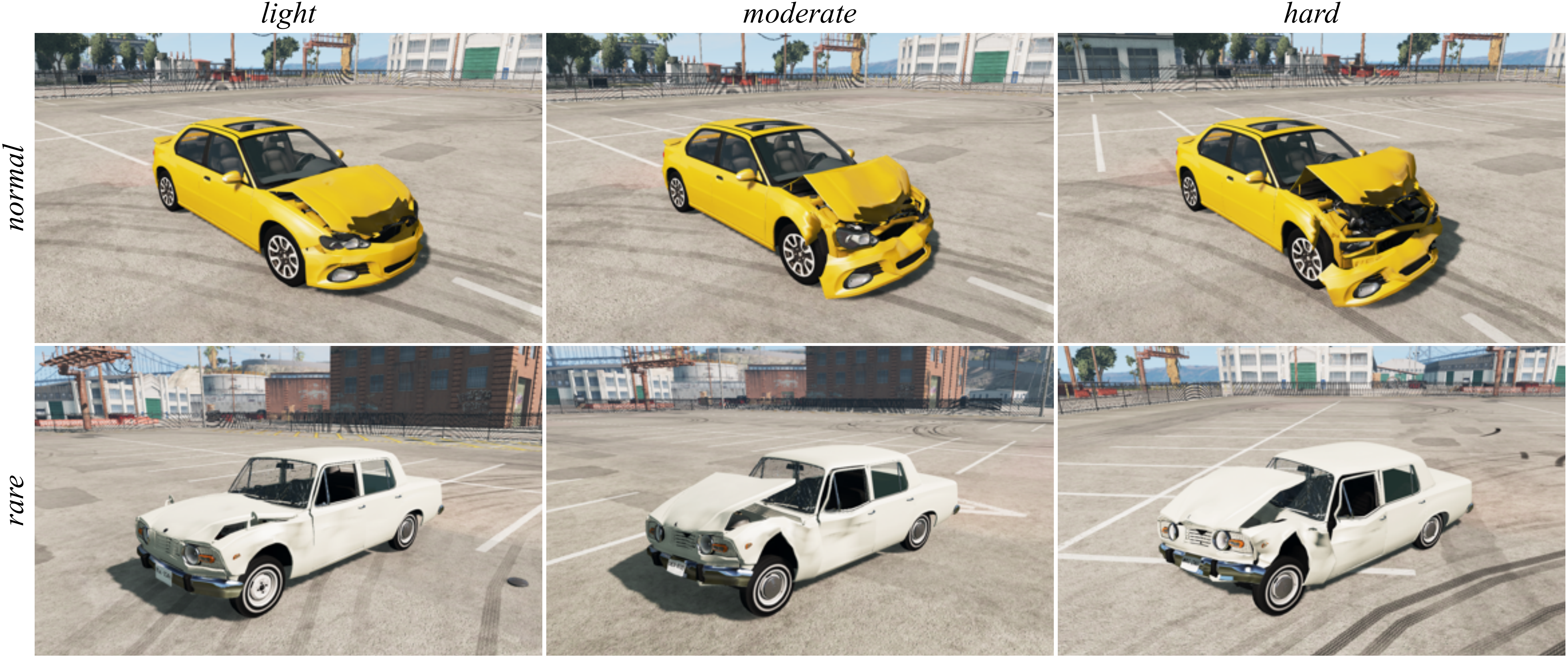}
   \caption{
        Comparison of \textit{linear} damage intensities for \textit{normal} and \textit{rare} cars of CrashD. For each type of car, the accidents were created by the same hitting vehicle, coming from the same angle. It can be seen that the \textit{hard} crash compromised the structure of the weaker \textit{rare} car, while the \textit{normal} car absorbed the impact differently, leaving the cabin unchanged.
   }
   \label{fig:intensity_comparison}
\end{figure*}

\subsubsection{Data Generation and Collection}
We generated random accidents with random settings (e.g., hitting angle, distance, type of hitting car, type of hit car), and placed the cars randomly in the LiDAR scene.
On each type of car (i.e., \textit{normal} and \textit{rare}), we applied 2 types of accidents (i.e., \textit{linear} and \textit{t-bone}), with 3 intensities each (i.e., \textit{light}, \textit{moderate} and \textit{hard}). That results in 12 different categories of damaged cars and their 12 undamaged counterparts (i.e., \textit{clean}), resulting in 24 categories overall. As the undamaged cars were placed at the exact same locations in the LiDAR scenes, they can be used as control group, to check the performance drop of a 3D detector when introducing the damages on the same cars.

We generated the accidents as follows. For each of the 12 categories of damages, we randomly selected 5 cars of the corresponding vehicle type (i.e., \textit{normal}, \textit{rare}), and 1 hitting vehicle. The hitting vehicle crashed into each of the 5 cars, getting repaired before each crash. We then repeated this process at least 64 times for each of the 12 categories, generating more than 3840 different accidents.

Furthermore, within each category, we used several random parameters, resulting in a high amount of possible damages. The intensity was determined by the distance from which the hitter starts, so the higher the distance, the higher the speed at which it will hit the target (i.e., one of the 5 cars). The effect of different intensities on the two types of cars for a \textit{linear} crash can be seen in Figure~\ref{fig:intensity_comparison}. For each intensity type, there was a random variable determining a variation of the distance at which the hitter was placed. Then, the hitting angle and the side (i.e., front or back for \textit{linear}, and left or right for \textit{t-bone}) were also randomized. Overall, this covered 360 degrees for each type of car and intensity.

Each batch of 5 cars, after being hit, was randomly placed in the LiDAR scene, such that the damaged area was visible from the sensor viewpoint, as shown in Figure~\ref{fig:lidar_scene}. We considered a crash visible if the sensor was within 35 degrees from the hitting angle. This ensured that a car classified as damaged is represented by a deformed point cloud. Moreover, if the damaged part was not visible from the sensor, the car was discarded from the batch.

This was due to a series of reasons, resulting in the lack of control over the rotation of the damaged car within the LiDAR scene.
In particular, BeamNG setup the simulator~\cite{BeamNGTechnicalPaper21} such that if a vehicle is rotated programmatically, it gets automatically repaired. Plus, depending on the dynamics of a crash, a damaged vehicle could rotate following the impact. So, as we reduced the LiDAR scene to the front 180 degrees, we had to discard some cars to be sure that they were not classified as damaged if their impacted area was not visible.
To avoid that crashes with a set of hitting angles could systematically not be placed in the scene, we randomly rotated the whole accident scenarios.

\begin{figure*}[t]
\centering
  \includegraphics[width=1.0\textwidth]{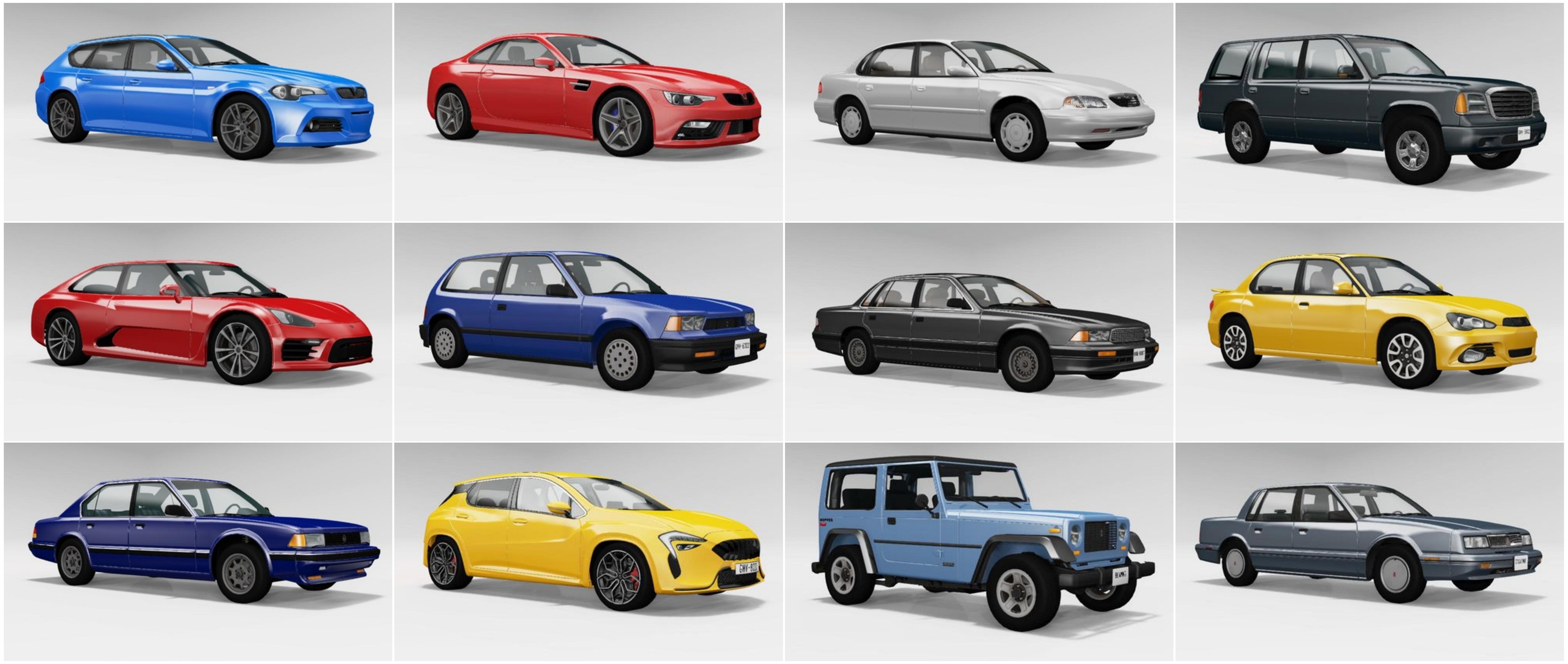}
   \caption{
        \textit{Normal} cars of CrashD. These were classified as \textit{normal} as they resemble the vast majority of cars on the road today in Germany, USA, and other locations where popular LiDAR datasets, such as KITTI~\cite{geiger2012we} and Waymo~\cite{sun2020waymo}, have been recorded.
   }
   \label{fig:normal_cars}
\end{figure*}

\begin{figure*}[t]
\centering
  \includegraphics[width=1.0\textwidth]{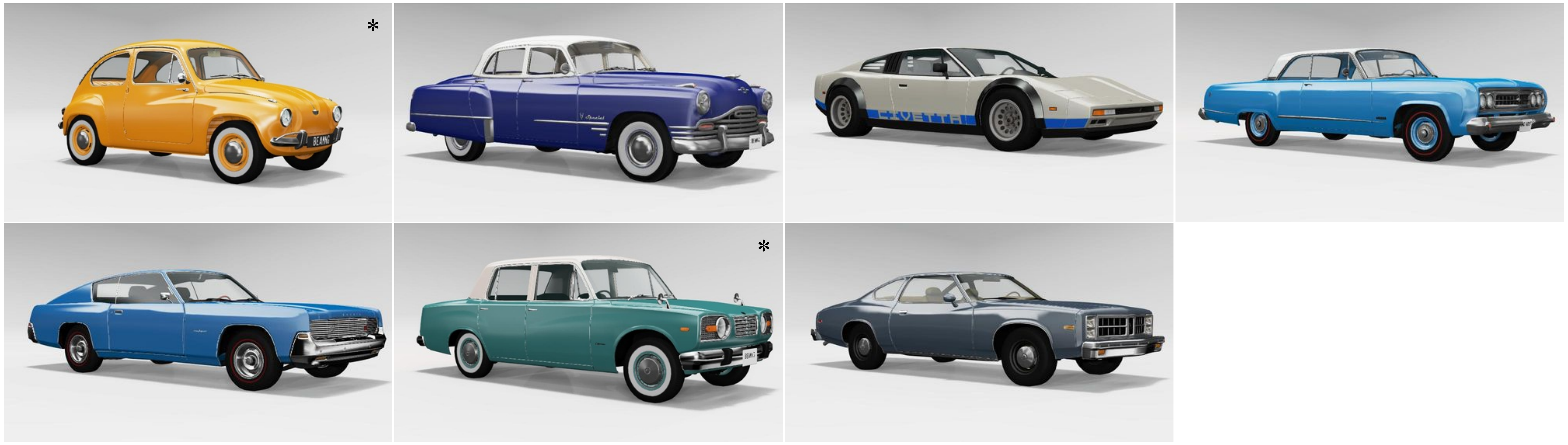}
   \caption{
        \textit{Rare} cars of CrashD. These were classified as \textit{rare} as they complement the \textit{normal} (i.e., common) cars shown in Figure~\ref{fig:normal_cars}. In particular, \textit{rare} ones resemble old cars from various regions, and also include a wedge-shaped sports car. * indicates cars that cannot hit other vehicles (due to their low speed and weight), but can only be hit by others.
   }
   \label{fig:rare_cars}
\end{figure*}

For each batch of 5 cars, we recorded 10 frames with the cars with visible damages (between 1 and 5), where we randomized the distance from the sensor, as well as the angle around it.
Moreover, again to avoid that a vehicle is considered damaged if the affected area is not visible, we excluded occlusions considering only 25 angles around the sensor, and preventing two cars from occupying the same one. This resulted in 750 possible different locations in the scene.
With this setup, a given vehicle might be discarded in one frame if the angles from which its damage is visible are occupied by other cars, but might appear in a subsequent frame if it gets placed beforehand.

Furthermore, we put the objects only in the front, motivated by the front-facing setup of KITTI~\cite{geiger2012we}, thereby facilitating transfers from KITTI to the proposed CrashD. Towards this end, we positioned the vehicles from 10 to 40 meters away from the LiDAR, around its front 180 degrees. As shown in Figure~\ref{fig:lidar_scene}, the scene features a large parking lot, where no object is located, other than the cars. We selected a totally empty parking lot (lacking poles, trees, or anything else), to fully focus on the task at hand, providing test data for evaluating the generalization capability of a method to different object shapes. Instead, having distracting elements (e.g., trees) in the scene, could have led to a different kind of transfer evaluation (e.g., the ability of recognizing cars compared to other objects in the scene), which goes beyond the scope of this dataset.
Nevertheless, in the main paper, as well as in additional results in this supplementary material, we also show a transfer from KITTI~\cite{geiger2012we} to Waymo~\cite{sun2020waymo}, which features real complex scenes, with trees and other objects, thereby challenging the 3D detector in a different way compared to transferring to the proposed CrashD.

\begin{figure*}[tb]
\centering
  \includegraphics[width=1.0\textwidth]{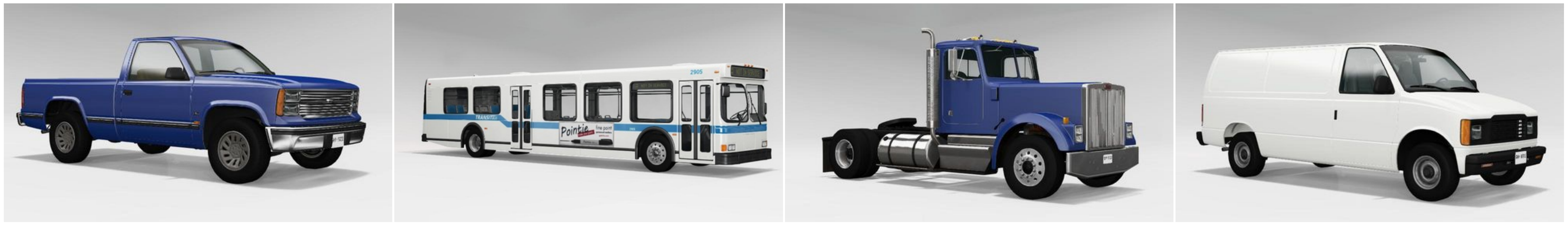}
   \caption{
        These vehicles can only hit others and are not detectable objects, as they do not fit the KITTI~\cite{geiger2012we} criteria for being a car, so they would not get recognized by a model transferred from KITTI.
   }
   \label{fig:hitter_cars}
\end{figure*}

\subsubsection{Vehicles}
The simulator offers a variety of fictional vehicles, which are shown in Figures~\ref{fig:normal_cars}, \ref{fig:rare_cars} and \ref{fig:hitter_cars}.
In particular, the 12 \textit{normal} cars used are shown in Figure~\ref{fig:normal_cars}, resembling the vast majority of vehicles on the road today in the countries where common LiDAR datasets were recorded, such as Germany and USA, for KITTI~\cite{geiger2012we} and Waymo~\cite{sun2020waymo} respectively.
Figure~\ref{fig:rare_cars} shows the 7 \textit{rare} cars used for CrashD, including older cars from Europe, USA and Asia, as well as a wedge-shaped sports car. Among older cars, the simulator features different muscle cars, and also a very small car (at the top left of Figure~\ref{fig:rare_cars}).

The significant gap between the two types of cars can be seen by comparing the \textit{normal} and \textit{rare} vehicles in Figures~\ref{fig:normal_cars} and \ref{fig:rare_cars} respectively.
Specifically, considering the \textit{normal} cars resemble those from KITTI and Waymo, the shapes of the \textit{rare} ones are rather different, posing a substantial challenge for any LiDAR-based 3D object detector transferring on this dataset from those two others. Analogously, detecting the cars with the various deformations resulting from the accidents, which can be seen in Figure~\ref{fig:intensity_comparison}, pose a different, but also significant challenge for a detector trained on KITTI, Waymo, or a similar dataset.

Since the KITTI~\cite{geiger2012we} \textit{car} annotations do not include vans, trucks, pickups and busses, we excluded these from the detectable vehicles of CrashD. Nevertheless, these vehicles were part of the pool of hitting vehicles, and they are shown in Figure~\ref{fig:hitter_cars}. Hitting vehicles also included all the ones shown in Figure~\ref{fig:normal_cars}, as well as those in Figure~\ref{fig:rare_cars}. However, we excluded the 2 cars marked with * due to their relatively low speed and weight, which would have not provided an accident as intense as those caused by the other vehicles, thereby altering the data distribution along the intensity types (i.e., \textit{light}, \textit{moderate}, \textit{hard}). In spite of that, the 2 with * were part of the detectable vehicles.

\subsubsection{Dataset Statistics}
In total, the proposed CrashD includes 46936 cars, half of which are damaged and half are not, as the LiDAR scenes were repeated with and without damages. \textit{Normal} cars are 23314, while \textit{rare} ones are 23622, again half of each is damaged. 8124 cars were hit by \textit{light} accidents, 7453 \textit{moderate} and 7891 \textit{hard}. 11530 were affected by a \textit{linear} crash, while 11938 by a \textit{t-bone}. Due to the vehicle placement in the LiDAR scene being dependent on the damage visibility, cars undergoing a \textit{linear} crash were more likely to be included from a frontal or rear perspective (including 3/4 views), while \textit{t-bone} ones were only included from the sides.

\subsection{Additional Implementation Details}\label{sec:appendix_impl}
\textbf{Iterative gradient L2 attack}
For this attack~\cite{xiang_generating_2019} we minimize our adversarial loss $\mathcal{L}_{\rm adv}$ constraining the deformation $\bm{m}$ for each point $\bm{p}$ with $\Vert \bm{m} \Vert_{2} < \epsilon$, with $\epsilon = 30$ cm.

\textbf{Chamfer attack}
For the Chamfer attack~\cite{liu2020adversarial} we used the Chamfer distance to measure the gap between the original and perturbed point clouds, which is given by:
\begin{equation}
\mathcal{C}(X, Y) = \frac{1}{|X|}\sum_{x \in X} \min_{y \in Y} ||x - y||_2
\end{equation}
for two sets $X$ and $Y$.
We perturb by minimizing:
\begin{equation}
\mathcal{L}_{\rm cha} = \mathcal{L}_{\rm adv} + \lambda \mathcal{C}(\bm{p} + \bm{m}, \bm{p})
\end{equation}
with $\lambda$ set to 0.1 and the amount of deformation constrained by $\mathcal{C}(\bm{p} + \bm{m}, \bm{p}) < \epsilon$, with $\epsilon = 30$ cm. It should be noted that single deformations vectors could lead to perturbations larger than 30 cm, since what is bounded is the overall Chamfer distance and not single vectors. This attack led to only a small amount of perturbed points, but the ones that moved showed large displacements.

\textbf{Adversarial removal}
For the removal attack we follow \cite{yang2019adversarial_removal} and remove 10\% of the \textit{critical points} of an object. These are those input points that if removed, the prediction changes. We estimate them as those with the highest deformation magnitude from the iterative gradient L2 attack~\cite{xiang_generating_2019}.

\textbf{Adversarial generation}
We follow \cite{xiang_generating_2019} adding 10\% of the objects points. We initialize their location as that of the \textit{critical points} (see removal). We then perform the iterative gradient L2 attack~\cite{xiang_generating_2019} solely on the added points. Thus shifting them to decrease the detection quality.

\textbf{Transfer to Waymo}
To evaluate the transfers to Waymo~\cite{sun2020waymo}, we used the standard KITTI evaluation. Therefore, the LiDAR scene was cut until 70 m in front of the ego vehicle and 40 m to both sides. We also lowered the whole point cloud and ground truth bounding boxes by 1.6 m, to match the KITTI coordinates and ground plane.

\subsection{Additional Outdoor Quantitative Results}\label{sec:appendix_quantitative}

\subsubsection{Transferability of the Vector Fields}
\begin{table}[h!]
\begin{center}
\begin{tabular}{l|cc|cc|cc}
\toprule
& \multicolumn{2}{c|}{PointP.~\cite{lang_pointpillars_2019}} &  \multicolumn{2}{c|}{Second~\cite{yan_second_2018}} &  \multicolumn{2}{c}{Part-A$^2$~\cite{shi2019parta2}} \\
Adv.aug. & AP & ASR & AP & ASR & AP & ASR \\
\midrule
none  & \textbf{77.1} & 63.4 & \textbf{79.2} & 54.9 & 79.2 & 50.5\\
{w/o} $\mathcal{L}_{adv}$ & 76.4 & 60.0 & 77.2 & 52.5 & \textbf{79.3} & 47.4\\

[ours] & \textbf{77.1} & \textbf{21.8} & 78.1 & \textbf{18.3} & \textbf{79.3} & \textbf{18.7}\\
\bottomrule
\end{tabular}
\end{center}
\caption{\textit{Moderate} AP and ASR $\downarrow$ across different models, showing transferability and efficacy of our deformations, on the validation set of KITTI. ASRs on Second and Part-A$^2$ are measured on vector fields trained on the defended PointPillars, to report the transferability. Adv.aug.: adversarial augmentation; {w/o} $\mathcal{L}_{adv}$: ours not learned.
}
\label{table:models}
\end{table}
Table~\ref{table:models} shows the high transferability of our adversarial deformations to other 3D object detectors. It can be seen that perturbations learned on PointPillars~\cite{lang_pointpillars_2019} are highly effective also on rather different architectures such as Second~\cite{yan_second_2018} and Part-A$^2$~\cite{shi2019parta2}, maintaining up to 86\% ASR across the models. Table~\ref{table:models} reports also the benefit of our adversarial augmentation strategy against our deformations. The perturbed point clouds targeting PointPillars are effective also to defend the other models.

\subsubsection{Robustness against noise} 
\begin{table}[h]
\begin{center}
\begin{tabular}{l|ccccc}
\toprule

Method & -10\% & -5\% & 0\% & +5\% & +10\% \\

\midrule

PointP.~\cite{lang_pointpillars_2019} & 70.51 & 70.88 & 77.11 & 67.36 & 65.27  \\

[ours] & \textbf{71.45} & \textbf{71.75} & \textbf{77.13} & \textbf{69.57} & \textbf{65.86} \\
\bottomrule
\end{tabular}
\end{center}
\caption{KITTI validation \textit{moderate} AP under various \% of removed and added points within the cars bounding boxes.}
\label{table:noise}
\end{table}
\begin{table*}[t]
\begin{center}
\begin{tabular}{c|c|l|ccc|ccc|ccc|ccc}
\toprule
\multicolumn{3}{l|}{} & \multicolumn{3}{c|}{\textit{normal, linear}} & \multicolumn{3}{c|}{\textit{normal, t-bone}} & \multicolumn{3}{c|}{\textit{rare, linear}} & \multicolumn{3}{c}{\textit{rare, t-bone}}\\
\multicolumn{3}{l|}{$\rightarrow$ CrashD} & \textit{light} & \textit{mod.} & \textit{hard} & \textit{light} & \textit{mod.} & \textit{hard} & \textit{light} & \textit{mod.} & \textit{hard} & \textit{light} & \textit{mod.} & \textit{hard}  \\

\midrule

\parbox[t]{2mm}{\multirow{4}{*}{\rotatebox[origin=c]{90}{PointP.~\cite{lang_pointpillars_2019}}}} & \multirow{2}{*}{clean} & baseline~\cite{lang_pointpillars_2019}  & 59.6 & \textbf{64.4} & 60.6 & 65.5 & 73.7 & 67.3 & 33.5 & 33.8 & 27.7 & 37.5 & 35.1 & 37.3 \\
&& 3D-VF [ours]  & \textbf{61.8} & 64.2 & \textbf{62.0} & \textbf{72.4} & \textbf{76.7} & \textbf{70.6} & \textbf{39.6} & \textbf{41.1} & \textbf{35.0} & \textbf{49.6} & \textbf{47.4} & \textbf{47.7}  \\

\cmidrule{2-15}

& \multirow{2}{*}{crash} & baseline~\cite{lang_pointpillars_2019} & 46.5 & 33.8 & 28.6 & 57.9 & 54.9 & 40.2 & 26.7 & 22.9 & 15.4 & 31.2 & 23.3 & 15.4 \\
&& 3D-VF [ours]  & \textbf{54.3} & \textbf{46.6} & \textbf{40.6} & \textbf{65.3} & \textbf{60.2} & \textbf{50.2} & \textbf{33.4} & \textbf{31.0} & \textbf{21.5} & \textbf{41.7} & \textbf{33.0} & \textbf{22.1} \\

\midrule

\parbox[t]{2mm}{\multirow{4}{*}{\rotatebox[origin=c]{90}{Second~\cite{yan_second_2018}}}}  & \multirow{2}{*}{clean} & baseline~\cite{yan_second_2018}  & 67.0 & 68.6 & 68.7 & 76.1 & 81.1 & 75.0 & 39.3 & 43.8 & 37.5 & 43.7 & 42.5 & 44.4\\
&& 3D-VF [ours] & \textbf{71.3} & \textbf{75.4} & \textbf{73.1} & \textbf{79.3} & \textbf{82.4} & \textbf{77.7} & \textbf{40.9} & \textbf{47.5} & \textbf{41.5} & \textbf{52.8} & \textbf{49.2} & \textbf{53.0}\\

\cmidrule{2-15}

& \multirow{2}{*}{crash} & baseline~\cite{yan_second_2018}  & 60.1 & 46.4 & 43.0 & 72.0 & 65.6 & 53.3 & 36.1 & \textbf{37.8} & 28.8 & 40.1 & 31.4 & 22.9\\
&& 3D-VF [ours]  & \textbf{64.8} & \textbf{50.4} & \textbf{44.9} & \textbf{75.5} & \textbf{69.4} & \textbf{58.1} & \textbf{38.4} & 37.0 & \textbf{29.1} & \textbf{49.3} & \textbf{37.7} & \textbf{25.4}\\

\midrule

\parbox[t]{2mm}{\multirow{4}{*}{\rotatebox[origin=c]{90}{Part-A$^2$~\cite{shi2019parta2}}}}  & \multirow{2}{*}{clean} & baseline~\cite{shi2019parta2} &
77.9& 82.7& 78.4& 86.6& 87.6& 85.2&  71.5& 72.7& 73.7& 78.3& 72.9& 75.1\\

&& 3D-VF [ours]& \textbf{85.6}& \textbf{86.2}& \textbf{86.0}& \textbf{91.3}& \textbf{93.2}& \textbf{90.5}&  \textbf{80.0}& \textbf{81.6}& \textbf{79.8}& \textbf{83.7}& \textbf{79.3}& \textbf{82.2}\\

\cmidrule{2-15}

& \multirow{2}{*}{crash} & baseline~\cite{shi2019parta2} & 71.1& 58.6& 49.3& 79.7& 64.3& 56.5& 61.7& 55.5& 49.0& 67.0& 48.6& 32.2\\
&& 3D-VF [ours] & \textbf{81.1}& \textbf{69.4}& \textbf{63.3}& \textbf{87.3}& \textbf{75.8}& \textbf{65.9}& \textbf{74.9}& \textbf{69.1}& \textbf{59.0}& \textbf{74.5}& \textbf{53.8}& \textbf{36.7}\\

\bottomrule
\end{tabular}
\end{center}
\caption{Detailed AP comparison of PointPillars~\cite{lang_pointpillars_2019}, Second~\cite{yan_second_2018}, and Part-A$^2$~\cite{shi2019parta2} trained on KITTI~\cite{geiger2012we} and transferred to the proposed CrashD without any fine-tuning. The evaluation is shown according to the various accident types, and intensities, as well as the kinds of car. Baseline indicates the standard method, while [ours] shows the impact of our adversarial augmentation strategy.}
\label{table:intensities_appendix}
\end{table*}
\begin{table*}[t]
\begin{center}

\begin{tabular}{l|r|rr|rr|rr}

\toprule

\multicolumn{2}{l|}{~} & \multicolumn{2}{c|}{IoU 0.1} & \multicolumn{2}{c|}{IoU 0.5} & \multicolumn{2}{c}{IoU 0.7} \\

\multicolumn{2}{l|}{$\rightarrow$ CrashD} & baseline~\cite{lang_pointpillars_2019} & 3D-VF [ours] & baseline~\cite{lang_pointpillars_2019} & 3D-VF [ours] & baseline~\cite{lang_pointpillars_2019} & 3D-VF [ours] \\

\midrule
\multirow{3}{*}{\textit{normal, clean}} & TP $\uparrow$ & 11547 &	\textbf{11651} & 11539 &	\textbf{11638} & 8571 &	\textbf{8894}\\
& FP  $\downarrow$ & 4069	& \textbf{419} & 4077&	\textbf{432} & 7045&	\textbf{3176} \\
& FN $\downarrow$ & 110	& \textbf{6} & 118	&\textbf{19} & 3086	&\textbf{2763} \\

\midrule
\multirow{3}{*}{\textit{normal, crash}} & TP $\uparrow$ &
        11485	& \textbf{11642} & 11391	& \textbf{11562} & 6770 &	\textbf{7620}\\
& FP $\downarrow$ & 4550	    & \textbf{772} &   4644	& \textbf{852}   & 9265 &	\textbf{4794}\\
& FN $\downarrow$ & 172	    & \textbf{15} &    266	&     \textbf{95}    & 4887 &	\textbf{4037}\\

\midrule
\multirow{3}{*}{\textit{rare, clean}} & TP $\uparrow$ & 11761 & \textbf{11805} &  11747 & \textbf{11790}  & 6091 & \textbf{7528}\\
& FP $\downarrow$ & 4700 & \textbf{316} & 4714 & \textbf{331} & 10370 & \textbf{4593}\\
& FN $\downarrow$ & 50 & \textbf{6}  & 64 & \textbf{21} &  5720 & \textbf{4283}\\

\midrule
\multirow{3}{*}{\textit{rare, crash}} & TP  $\uparrow$ & 11724 & \textbf{11804}  & 11566 & \textbf{11680} &  4688 & \textbf{6011}\\
& FP $\downarrow$ & 4742 & \textbf{590} & 4900 & \textbf{714} & 11778 & \textbf{6383}\\
& FN $\downarrow$ & 87 & \textbf{7}   & 245 & \textbf{131} & 7123 & \textbf{5800}\\

\bottomrule

\end{tabular}

%
%
%
%
%
\end{center}
\caption{
    Impact of our adversarial augmentation on the main categories of the proposed CrashD according to true positives (TP), false positives (FP) and false negatives (FN) at different IoU thresholds. The models were based on PointPillars~\cite{lang_pointpillars_2019}, trained on KITTI~\cite{geiger2012we} and transferred to CrashD without any fine-tuning. For reference, the total amount of cars in CrashD is 46936.
}
\label{table:all_crashD_tpfpfn}
\end{table*}
In Table~\ref{table:noise} we report the performance of PointPillars~\cite{lang_pointpillars_2019} with and without our adversarial augmentation strategy.
For this set of experiments, at inference time we randomly added and removed points within the cars bounding boxes according to the percentages reported in the table. Both models were the same as in the rest of this work, simply evaluated with this setup.
Thanks to the improved generalization provided by our vector fields, the augmented model was more robust against such noise. Our augmentation acts as regularization during training, allowing the model to learn more meaningful features independent of specific points. This led to a constant gap between 5 and 10\% removal. Conversely, randomly adding points is not realistic from the sensor perspective, since occlusions and its physical properties are not respected. Due to this reason, both models suffered more when adding points, than removing.

\subsubsection{Detailed transfer to CrashD}
\textbf{Evaluation by categories} In Table~\ref{table:intensities_appendix}, we show a detailed evaluation of the various 3D object detectors along the different sub-categories of the proposed CrashD, with various kinds of damages, different intensities and types of cars.
Our adversarial augmentation strategy outperformed all detectors~\cite{lang_pointpillars_2019,yan_second_2018,shi2019parta2} across the board by a significant margin, especially on \textit{rare} cars.
In particular, with high intensity crashes (\textit{hard}), the baselines~\cite{lang_pointpillars_2019,yan_second_2018,shi2019parta2} severely underperformed, reducing by half their APs on cars undergoing a \textit{t-bone} accident.
This can be due to the large point displacement introduced by the impacts, especially with weaker old cars.
Conversely, our 3D-VField, as it was trained on sensor-aware deformations,
was more robust against these damages, delivering a smaller decrease from the \textit{clean} cars to their \textit{crash} counterparts.
Interestingly, \textit{rare} vehicles were often more challenging to be detected than \textit{normal crash} ones. This can be attributed to an accident typically affecting only a local region of a vehicle, leaving the rest of it untouched and detectable, compared to a rare design which has an impact on the whole object point cloud, making it in general harder to be recognized.
Comparing the same cars with and without damages (\textit{crash} and \textit{clean}) shows that the former are significantly more difficult for every detector, due to the different resulting shapes.
All detectors substantially benefited from our adversarial augmentations, despite training the vector fields solely against PointPillars~\cite{lang_pointpillars_2019}. The values also confirm the superiority of Part-A$^2$~\cite{shi2019parta2} over the other 3D detectors, as seen in Table~\ref{table:AP_on_all}. 

\textbf{Correct and wrong detections on CrashD}
Table~\ref{table:all_crashD_tpfpfn} reports a comparison of PointPillars~\cite{lang_pointpillars_2019} without and with our adversarial augmentations on CrashD, according to the number of true positives, false positives and false negatives, depending on the main categories of the proposed dataset, at different IoU thresholds.
It can be seen that the baseline~\cite{lang_pointpillars_2019} had a strong tendency towards over-predicting the amount of objects in the scene, resulting in a high number of false positives. In fact, even with a low IoU threshold of 0.1, over 30\% of the boxes predicted by the baseline did not match any car in the scene.
At the same time, it completely ignored several cars, both damaged and undamaged, resulting in false negatives.
On the other hand, as seen already in the main paper showing the APs, the proposed 3D-VField delivered a significantly better detection rate, vastly reducing the amount of false positives and negatives, despite being based on the same architecture and settings as the baseline~\cite{lang_pointpillars_2019}.

\subsubsection{Ablation Studies}

\begin{table}[h]
\setlength{\tabcolsep}{5.75pt}
\begin{center}
\begin{tabular}{l|c|c|cc}
\toprule
 & \multicolumn{1}{c|}{KITTI} & \multicolumn{1}{c|}{$\rightarrow$ W.} & \multicolumn{2}{c}{$\rightarrow$ CrashD} \\

\# augm.~objects & \multicolumn{1}{c|}{\textit{mod.}} & & \textit{n.,clean} & \textit{r.,crash} \\

\midrule

[ours] 1 obj. & \textbf{77.13} & \textbf{44.61} & \textbf{67.95} & \textbf{30.37} \\

[ours] 50\% obj. & 76.31 & 39.60 & 53.99 & 23.63\\

[ours] 100\% obj. & 59.30 & 32.84 & 38.29 & 14.83\\

\bottomrule
\end{tabular}
\end{center}
\caption{Models trained on KITTI, augmented with our adversarial technique. In each row, the amount of objects augmented at training time in each scene changes. The chosen number of augmented objects was 1. \textit{mod.}: moderate difficulty; $\rightarrow$: transfer without any fine-tuning; W.: Waymo; \textit{n.}: \textit{normal}; \textit{r.}: \textit{rare}.}
\label{table:ablation_num_objects}
\end{table}
\textbf{Amount of deformed objects}
In Table~\ref{table:ablation_num_objects} we report the effect of augmenting various amounts of objects during training. Specifically, augmenting more objects in each scene did not help generalization, as it made difficult to recognize standard objects. Augmenting all cars means the detector never learns a normal vehicle, making it rather hard to identify one at inference time. This can be seen in the AP drop on KITTI~\cite{geiger2012we} from augmenting half of the cars, to all of them. Instead, augmenting a single object allowed to retain the same AP on KITTI, while significantly improving it on the out-of-domain Waymo~\cite{sun2020waymo} and the proposed CrashD.

\textbf{Grouping strategies}
In Table~\ref{table:ablation_groupings} we show the impact of varying amounts of learned vector fields on the ASR, according to different distinguishing criteria. We compare the chosen relative rotation (Section~\ref{sec:field_application}) with selecting by distance of the object to the sensor or number of object points. Relative rotation delivered superior ASR, as it favors the mutual alignment between neighboring vectors. In contrast, less vector fields (i.e., 1 and 6) or different criteria resulted in contrasting vectors, reducing the object deformation.

\begin{table}[h!]
\begin{center}
\begin{tabular}{l|cccc}
\toprule
Grouping & 1-ASR & 6-ASR & 12-ASR & 18-ASR \\
\midrule
distance & \textbf{55.1} & 56.2 & 57.3 & 57.6 \\
nr. points & \textbf{55.1} & 56.9 & 56.0 & 57.1\\
rel. rotation & \textbf{55.1} & \textbf{59.2} & \textbf{63.4} & \textbf{63.7} \\
\bottomrule
\end{tabular}
\end{center}
\caption{ASR $\uparrow$ on the validation set of KITTI for different grouping strategies and amount of vector fields.}
\label{table:ablation_groupings}
\end{table}

\textbf{Aggregation strategies}
Table~\ref{table:ablation_aggregation} shows the effect of different aggregation strategies of vectors when applying the deformations on the cars of KITTI~\cite{geiger2012we}. It can be seen how the different amount of groupings (G) and neighboring vectors (k) considered for each point shift affected the adversarial performance of the method (ASR).
In general, all deformations in the table were restricted to a maximum of $\epsilon =30$ cm. The amount of learned vector fields $G$ had an impact on the ASR of each aggregation strategy. For example, sum was more effective with 12 $G$ than 1 $G$, since the vectors of the 12 fields were better aligned to each other than those of the single field (Section~\ref{sec:quantitative}), so summing them increased the deformation magnitude. In fact, the high ASR of sum with 12 $G$, was due to larger perturbations.

\setlength{\tabcolsep}{5pt}
\begin{table}[t]
\begin{center}
\begin{tabular}{l|c|cc|cc|cc}
\toprule
 & \multicolumn{7}{c}{Aggregation} \\
& \multicolumn{1}{c|}{-} & \multicolumn{2}{c|}{sum} & \multicolumn{2}{c|}{average} & \multicolumn{2}{c}{distance} \\
\multicolumn{1}{r|}{$k$}& 1 & 2 &3 & 2 & 3 & 2 &3 \\
\midrule
\multirow{1}{*}{$G=1$} & 46.3 & 44.4 & 33.4& 45.4& \textbf{52.0}& 50.3& 47.0\\
\midrule
\multirow{1}{*}{$G=12$} & 59.6  & 76.5& \textbf{80.3} & 61.9 & 62.4 & 63.4 & 59.6 \\
\bottomrule
\end{tabular}
\end{center}
\caption{ASR $\uparrow$ on the validation set of KITTI~\cite{geiger2012we} for different aggregation strategies and number of neighbors ($k$) involved in each deformation, for both number of groups $G=1$ and $G=12$. All configurations are based on PointPillars~\cite{lang_pointpillars_2019}.}
\label{table:ablation_aggregation}
\end{table}

\textbf{Grid step size}
In Table~\ref{table:ablation_step} we show the impact of different step sizes $t$ of the vector field grid. A larger step size, results in a coarser grid, which in turn means less vectors for each field. Intuitively, with more vectors, each would be more specific for a given point shift, but less generalizable to others. So, each vector would overfit to its training points.
There is in fact a trade-off between the amount of vectors and the generalizability of the learned vector field, as seen in Table~\ref{table:relative_rotations}.
That can be seen by the ASR, as the vectors were learned on the training set of KITTI~\cite{geiger2012we}, and applied to its validation set, on which the values are reported.
Down-scaling $t$ from 20 to 5 cm, significantly reduced the ASR. Conversely, increasing $t$ to 30 cm worsened their generalization.
Therefore, $t=20$ cm was chosen as the grid step size, offering a good trade-off between the vector specificity and generalizability, as shown by the ASR.

\begin{table}[h]
\begin{center}
\begin{tabular}{ll|cccc}
\toprule
\multicolumn{2}{l|}{Step size} & 5 cm & 10 cm & 20 cm & 30 cm\\
\midrule
\multicolumn{2}{r|}{ASR $\uparrow$} &   44.1  &  46.3 &  \textbf{53.0} &  49.6 \\
\bottomrule
\end{tabular}
\end{center}
\caption{ASR $\uparrow$ on the validation set of KITTI~\cite{geiger2012we} for different step sizes of the vector field grid. A smaller step size increases the amount of vectors. All configurations are based on PointPillars~\cite{lang_pointpillars_2019}, with $G=1$.}
\label{table:ablation_step}
\end{table}

\subsection{Additional Outdoor Qualitative Results}
\label{sec:appendix_qualitative}
In this section we provide qualitative results of the learned deformations.

\subsubsection{Deformations on KITTI}
\begin{figure}[t]
    \begin{center}
        \includegraphics[width=1.\linewidth]{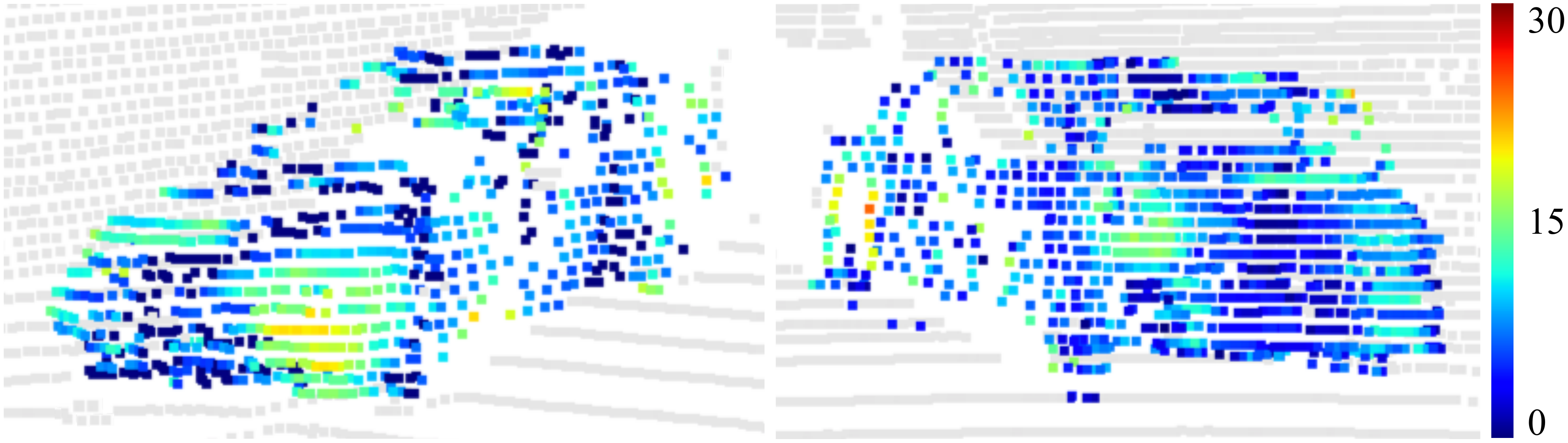}
    \end{center}
    \caption{Color-coded deformations in cm learned by the proposed method. The perturbation does not affect every point, its magnitude is relatively low, and local smoothness is preserved.}
    \label{fig:dent}
\end{figure}
Figure~\ref{fig:dent} shows the deformations learned by our method. It can be seen that only local areas are affected, and the cars preserved their overall shapes with smoothly deformed parts.

\begin{figure}[h]
\begin{center}
\includegraphics[width=1.\linewidth]{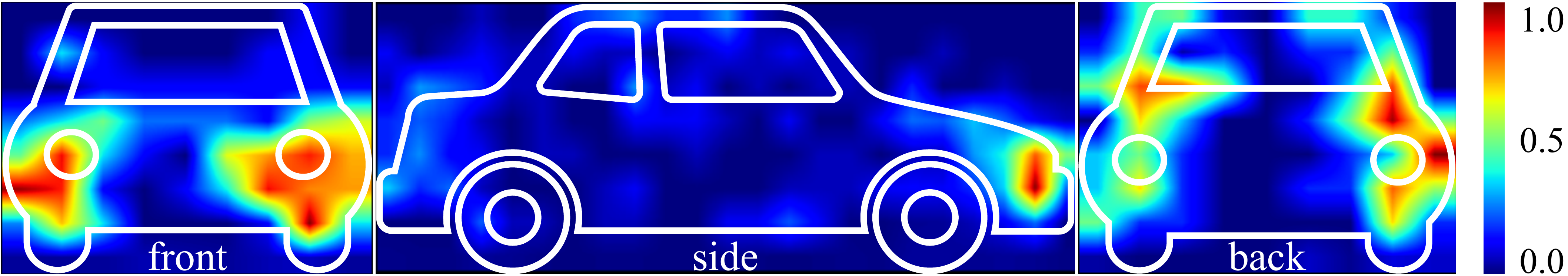}
\end{center}
   \caption{Color-coded contribution of each vector to the ASR, in percentage.}
\label{fig:realism}
\end{figure}

Figure~\ref{fig:realism} shows the effect of each vector of the adversarial field to the ASR. It can be seen that the most affected was the front bumper, which can easily be deformed with an accident. The side of the car is mostly unaffected, probably due to the relatively limited amount of vehicles visible from the side in KITTI.
Interestingly, the model has learned to avoid the areas without points (e.g., the windows).

\begin{figure*}[t!]
\centering
  \includegraphics[width=1.0\textwidth]{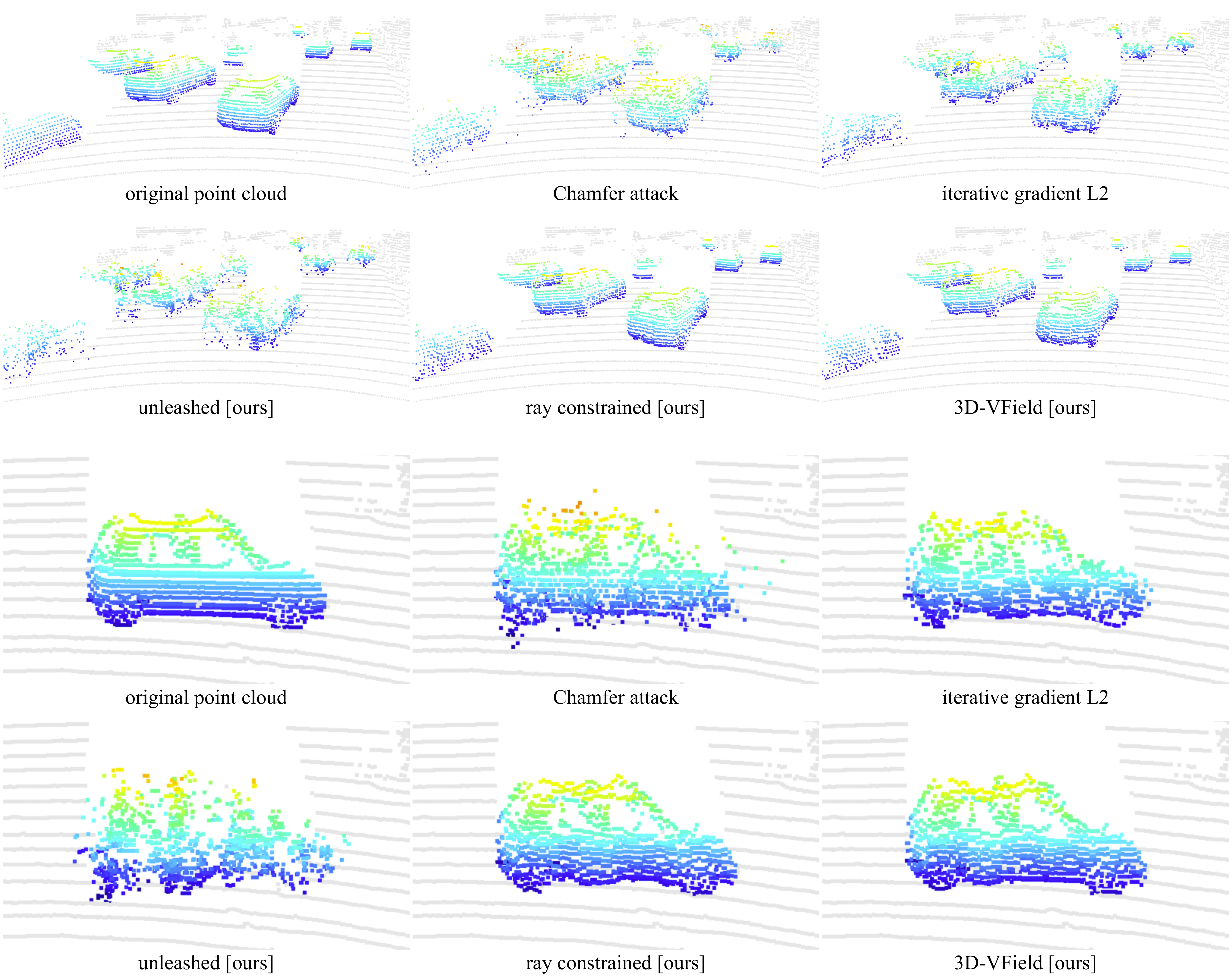}
   \caption{
        Comparison of adversarial perturbations on a set of cars from two different point clouds of KITTI~\cite{geiger2012we}. The effect of the Chamfer attack~\cite{liu2020adversarial}, the iterative gradient L2~\cite{xiang_generating_2019}, and multiple variations of our approach are shown. It can be seen that our 3D-VField preserves the shape of the original point cloud better than the other approaches.
   }
   \label{fig:deformations_comp}
\end{figure*}

Figure~\ref{fig:deformations_comp} shows a comparison of the deformations applied by each method to a set of cars from KITTI~\cite{geiger2012we}.
We included related works, such as the Chamfer attack~\cite{liu2020adversarial} and the iterative gradient L2 approach~\cite{xiang_generating_2019}, as well as variations of the proposed 3D-VField.
The Chamfer attack~\cite{liu2020adversarial} shifted some points far away while many remained close to the original location, resulting in an almost perfect ASR. However, this came at the cost of rather obvious perturbations. The iterative gradient L2~\cite{xiang_generating_2019} method also achieved a highly effective ASR (Table~\ref{table:AP_on_all}), but with significantly less evident deformations.
As expected from the high ASR (Table~\ref{table:ablation_learning}), our unconstrained (unleashed) method delivered substantially perturbed objects, even more distorted than those produced by the Chamfer attack. 
Applying the ray constraint allowed for less perturbed (and less effective ASR), but more recognizable objects. It can be seen how this constraint alone impacts the realism of the deformations, by comparing it to the unleashed version.
Moreover, aggregating neighboring vectors via distance weighting in our full approach (Section~\ref{sec:field_application}), further improved the resemblance of the object to the original point cloud. Although the difference is subtle, this can be appreciated comparing the rear wheel, the floor, and the windows of the car in the bottom half of Figure~\ref{fig:deformations_comp}. 
Thanks to the realism and the smooth alterations of the points visible in the figure, training with our deformations allowed for superior transfer performance to challenging out-of-domain data (Table~\ref{table:AP_on_all}).

\subsection{Results on Indoor Data}\label{sec:indoor_appendix}

\subsubsection{Experimental setup}
In these experiments we used the SUN RGB-D dataset~\cite{song2015sun}, which posed a completely new set of challenges compared to the three driving datasets. SUN RGB-D contains indoor furniture objects captured by depth cameras such as time-of-flight (ToF), as opposed to driving scenes captured by a LiDAR. We trained on all 10 classes, but we selected one at a time for learning our vector fields. In particular, we report on the classes \textit{bed}, \textit{sofa} and the highly diverse \textit{chair}, as they are the ones where deformations are more plausible compared to others (e.g., \textit{table}). In this setting, we apply our method on a VoteNet~\cite{qi2019_votenet} architecture. Moreover, we followed the same setup as for the outdoor experiments, except that we reduced the maximum deformation $\epsilon$ to 10 cm, making it more plausible in indoor settings.

\subsubsection{Quantitative Results}
\begin{table}[h]
\begin{center}
\begin{tabular}{l|cc|cc|cc}
\toprule
& \multicolumn{2}{c|}{\textit{beds}} &  \multicolumn{2}{c|}{\textit{sofas}} &  \multicolumn{2}{c}{\textit{chairs}} \\
Adv.aug. & AP & ASR & AP & ASR & AP & ASR \\
\midrule
none  & 85.6 & 49.7 & 67.4 & 70.6 &  77.4 & 70.9\\
{w/o} $\mathcal{L}_{adv}$  & 85.2 & 41.1 & 67.5 & 65.4 & 76.9 & 62.1\\

[ours]  & \textbf{86.0} & \textbf{19.7} & \textbf{68.5} & \textbf{34.8} & \textbf{77.5} & \textbf{39.6} \\
\bottomrule
\end{tabular}
\end{center}
\caption{AP and ASR $\downarrow$ on the validation set of SUN RGB-D~\cite{song2015sun}, with a VoteNet~\cite{qi2019_votenet} architecture. Adv.aug.: adversarial augmentation; {w/o} $\mathcal{L}_{adv}$: ours not learned.}
\label{table:sunrgbd}
\end{table}
Table~\ref{table:sunrgbd} shows the wide applicability of our deformation and augmentation strategies when applied to point clouds from depth sensors capturing furniture objects from SUN RGB-D~\cite{song2015sun}.
Shifting the points with our 3D-VField produced a strong ASR against the not adversarially augmented models (none), especially on \textit{sofas} and \textit{chairs}. Using the deformations as augmentation even improved the AP on the validation set, confirming the benefit of our techniques towards the generalization to unseen data, despite the rather different setting, sensor, objects, and architecture. Furthermore, defending with our adversarial augmentations significantly reduced the ASR, showing the gained robustness against deformed objects.

\begin{figure}[h]
\begin{center}
\includegraphics[width=1.00\linewidth]{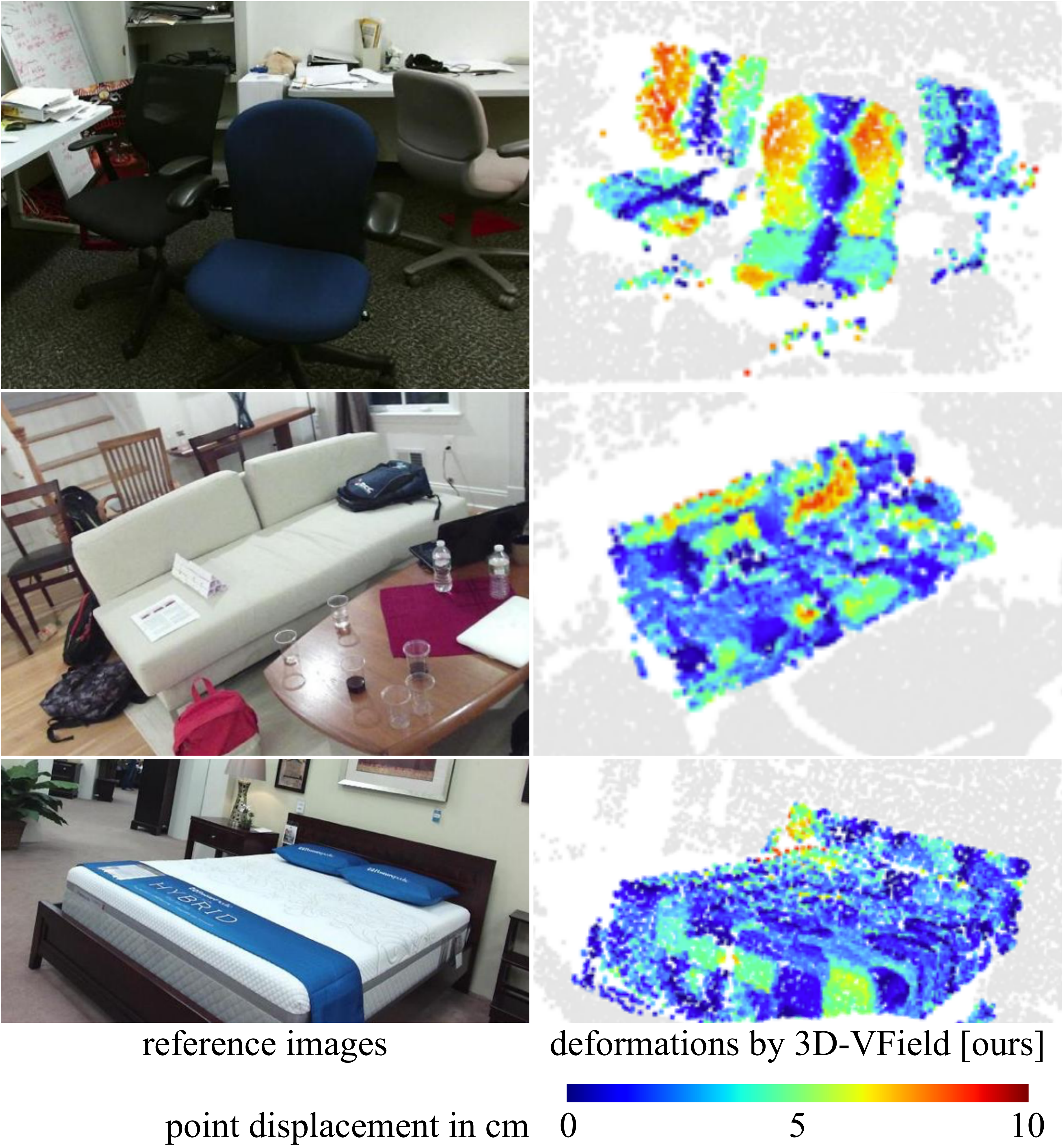}
\end{center}
   \caption{Color-coded deformations applied by the proposed 3D-VField on various objects of the SUN RGB-D dataset~\cite{song2015sun}. The color corresponds to the shift of each point in centimeters, limited to a maximum of 10 cm. Adversarial deformations learned against VoteNet~\cite{qi2019_votenet}.}
\label{fig:deformations_sun}
\end{figure}

\subsubsection{Qualitative Results}
In Figure~\ref{fig:deformations_sun} we show the deformations learned by our method against VoteNet~\cite{qi2019_votenet} on three different categories of objects from SUN RGB-D~\cite{song2015sun}, namely \textit{chairs}, \textit{sofas}, and \textit{beds}. It can be seen that the overall shape of each object is preserved, with minor perturbations applied. In this indoor setting, such alterations could resemble the presence of pillows, a blanket, or simply a different design of the object.

{\small
\bibliographystyle{ieee_fullname}
\bibliography{bib}
}


\end{document}